\documentclass[11pt, a4paper, logo, copyright, nonumbering]{antgroup}
\usepackage[square, numbers]{natbib}
\usepackage{dblfloatfix}
\usepackage{ulem}
\usepackage{caption}
\usepackage{dramatist}
\usepackage{xspace}
\usepackage{pifont}
\usepackage{multirow}
\usepackage[ruled,vlined]{algorithm2e}

\usepackage{xltabular}
\usepackage{longtable}
\usepackage{hyperref}
\usepackage{amsfonts}
\usepackage{amsmath}
\usepackage{amssymb}
\usepackage{lineno}
\usepackage{multirow}
\usepackage{adjustbox}

\usepackage[bottom]{footmisc}

\usepackage{CJKutf8}
\usepackage{subfigure}
\usepackage{setspace}

\usepackage{dsfont}
\usepackage{array}
\usepackage{tabularx}
\usepackage{subfigure}
\usepackage{xcolor}
\usepackage{tabularx}
\usepackage{booktabs}

\usepackage{lipsum}
\usepackage{multicol}
\usepackage{authblk}
\usepackage{wrapfig}
\usepackage[most,skins,theorems]{tcolorbox}
\usepackage{array}
\usepackage{siunitx}
\usepackage{pifont}
\definecolor{customgray}{RGB}{230,230,230}
\usepackage{algorithmic}

\tcbset{
  aibox/.style={
    width=\linewidth,
    top=8pt,
    bottom=4pt,
    colback=gray!10!white,
    colframe=gray!50!black,
    colbacktitle=gray!70!black,
    enhanced,
    center,
    attach boxed title to top left={yshift=-0.1in,xshift=0.15in},
    boxed title style={boxrule=0pt,colframe=white,},
  }
}
\newtcolorbox{AIbox}[2][]{aibox,title=#2,#1}

\makeatletter
\def\@BTrule[#1]{%
  \ifx\longtable\undefined
    \let\@BTswitch\@BTnormal
  \else\ifx\hline\LT@hline
    \nobreak
    \let\@BTswitch\@BLTrule
  \else
     \let\@BTswitch\@BTnormal
  \fi\fi
  \global\@thisrulewidth=#1\relax
  \ifnum\@thisruleclass=\tw@\vskip\@aboverulesep\else
  \ifnum\@lastruleclass=\z@\vskip\@aboverulesep\else
  \ifnum\@lastruleclass=\@ne\vskip\doublerulesep\fi\fi\fi
  \@BTswitch}
\makeatother

\addto\extrasenglish{
}

 {\begin{list}{}%
         {\setlength{\leftmargin}{#1}}%
         \item[]%
 }
 {\end{list}}

\reportnumber{001} %

\renewcommand{\today}{}

\title{\centering Agentar-Fin-R1: Enhancing Financial Intelligence through Domain Expertise, Training Efficiency, and Advanced Reasoning}

\author{
\centering
Yanjun Zheng$^{*}$,
Xiyang Du$^{*}$,
Longfei Liao$^{*}$
\par
\vspace{1pt}
\centering
Xiaoke Zhao,
Zhaowen Zhou,
Jingze Song,
Bo Zhang,
Jiawei Liu
\par
\vspace{2pt}
\centering
Xiang Qi,
Zhe Li,
Zhiqiang Zhang,
Wei Wang,
Peng Zhang$^{\dag}$
\par
\vspace{7pt}
Ant Group
}

\footnotetext[1]{Equal contribution.}
\footnotetext[2]{Corresponding Author.}
\footnotetext[3]{\{zhengyanjun.zyj, duxiyang.dxy, liaolongfei.llf\}@antgroup.com}

\renewcommand{\phi}{\varphi}

\renewcommand{\geq}{\geqslant}

\renewcommand{\epsilon}{\varepsilon}
\renewcommand{\imath}{\mathrm{i}}

\newlength{\restsubwidth}
\newlength{\restsubheight}
\newlength{\restsubmoreheight}
\newcommand{\rest}[2]{%
        \settowidth{\restsubwidth}{\ensuremath{#2}}
        \settoheight{\restsubheight}{\ensuremath{{}_{#2}}}
        \ensuremath{{#1\hskip 0.5pt}_{\vrule\kern2pt\parbox[b][%
        4pt][b]{\the\restsubwidth}{%
                        \ensuremath{{}_{#2}}}}}
        }

\begin{abstract}
Large Language Models (LLMs) exhibit considerable promise in financial applications; however, prevailing models frequently demonstrate limitations when confronted with scenarios that necessitate sophisticated reasoning capabilities, stringent trustworthiness criteria, and efficient adaptation to domain-specific requirements. We introduce the \textbf{Agentar-Fin-R1} series of financial large language models (8B and 32B parameters), specifically engineered based on the Qwen3 foundation model to enhance reasoning capabilities, reliability, and domain specialization for financial applications. Our optimization approach integrates a high-quality, systematic financial task label system with a comprehensive multi-layered trustworthiness assurance framework. This framework encompasses high-quality trustworthy knowledge engineering, multi-agent trustworthy data synthesis, and rigorous data validation governance. Through label-guided automated difficulty-aware optimization, tow-stage training pipeline, and dynamic attribution systems, we achieve substantial improvements in training efficiency. Our models undergo comprehensive evaluation on mainstream financial benchmarks including FinEval 1.0, and FinanceIQ, as well as general reasoning datasets such as MATH-500 and GPQA-diamond. To thoroughly assess real-world deployment capabilities, we innovatively propose the \textbf{Finova} evaluation benchmark, which focuses on agent-level financial reasoning and compliance verification. Experimental results demonstrate that Agentar-Fin-R1 not only achieves state-of-the-art performance on financial tasks but also exhibits exceptional general reasoning capabilities, validating its effectiveness as a trustworthy solution for high-stakes financial applications. The \textbf{Finova} bench is available at \url{https://github.com/antgroup/Finova}.

\end{abstract}

\begin{document}
\maketitle
\clearpage
\section{Introduction}

\begin{figure*}[!htb] 
  \centering    
  \setlength{\abovecaptionskip}{1pt} 
  \setlength{\belowcaptionskip}{1pt} 
  \includegraphics[width=0.95\textwidth]{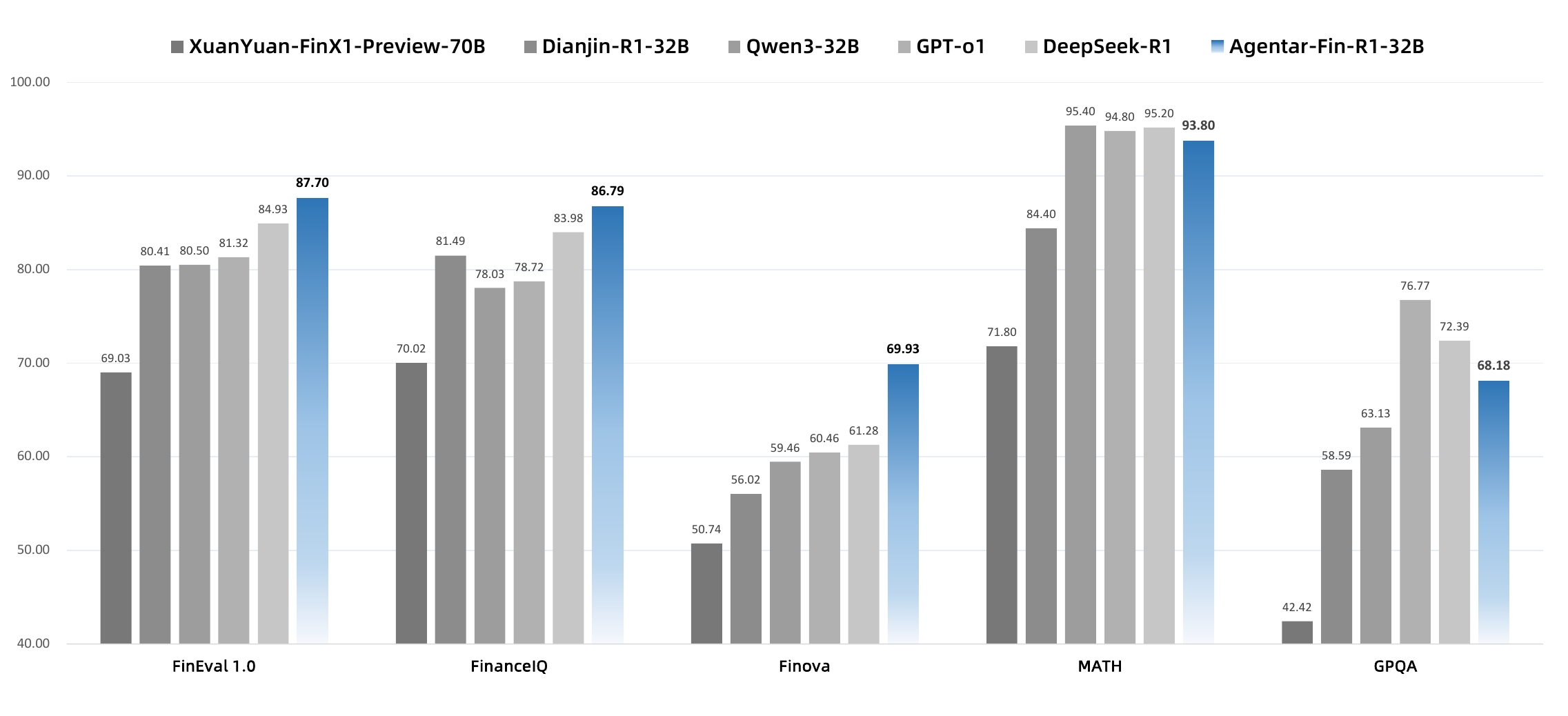} 
  \caption{Agentar-Fin-R1-32B performance: significantly outperforms on financial benchmarks (FinEval 1.0, FinanceIQ, Finova), and on general reasoning benchmarks (MATH-500, GPQA-diamond), outperforms comparable-sized general models while achieving performance comparable to large-scale models like DeepSeek R1 and GPT-o1.}
  \label{main}
\end{figure*}
Large Language Models (LLMs) have demonstrated remarkable capabilities in complex reasoning tasks, with recent advances in reasoning-optimized models such as OpenAI's o1 series~\citep{openai-2024-reasoning}, QwQ~\citep{qwen-2024-qwq}, DeepSeek-R1~\citep{guo-etal-2025-deepseek-r1},Seed-Thinking-v1.5\citep{seed2025seed1} and Qwen3~\citep{yang2025qwen3technicalreport} achieving significant breakthroughs in mathematics, programming, and logical inference. However, the direct deployment of these general-purpose models in financial applications reveals critical limitations:  insufficient domain-specific financial knowledge integration, leading to poor performance on finance-related tasks; susceptibility to hallucinations that violate the stringent safety and compliance requirements essential in financial environments.

Existing financial LLMs can be broadly categorized into two types. \textit{Non-reasoning financial models} such as Baichuan~\citep{zhang2025baichuan4financetechnicalreport}, DISC-FinLLM\citep{chen2023discfinllmchinesefinanciallarge}, XuanYuan~\citep{xuanyuan-72b}, and PIXIU~\citep{xie2023pixiulargelanguagemodel} incorporate domain-specific financial knowledge but lack sophisticated analytical and reasoning capabilities required for complex financial decision-making scenarios involving multi-step analysis, risk assessment, and strategic planning. \textit{Reasoning-enhanced financial models} including XuanYuan-FinX1-Preview~\citep{xuanyuan2024}, Fino1~\citep{qian-etal-2025-fino1}, Fin-R1~\citep{liu-etal-2025-finr1}, and Dianjin-R1~\citep{zhu2025dianjinr1evaluatingenhancingfinancial} attempt to integrate advanced reasoning mechanisms but still exhibit significant limitations: insufficient reasoning capabilities for handling complex financial scenarios that require deep analytical thinking; lack of scenario-specific reasoning adaptation, failing to align reasoning processes with the unique demands of financial contexts such as market dynamics, regulatory compliance constraints, and risk tolerance considerations.

We refer to some current consensus in academia and industry(\citet{liu-etal-2025-finr1},\citet{wang2024quantagent},\citet{dong2024fnspid},\citet{fatouros2024can},\citet{li2024alphafin},\citet{tong2024ploutos},\citet{xie2024finnlp},\citet{zhang2025baichuan4financetechnicalreport}), and here identify three fundamental requirements for effective financial AI systems that distinguish them from general-purpose applications:

\begin{enumerate}
    \item \textbf{Adaptive Knowledge Integration}: Efficient acquisition and assimilation of evolving domain knowledge, including regulatory updates and emerging financial instruments.
    \item \textbf{Verifiable Reasoning}: Transparent, auditable reasoning processes essential for stakeholder confidence in high-stakes decisions.
    \item \textbf{Compliance Adherence}: Robust protection of sensitive data while meeting stringent regulatory standards.
\end{enumerate}

Based on the aforementioned limitations, we introduce \textbf{Agentar-Fin-R1}, a family of reasoning-optimized financial LLMs that systematically addresses key challenges in the financial domain through three core innovations:

\textbf{Professional Label-Guided Framework}: We construct a fine-grained financial task label system that decomposes the financial domain into precisely defined categories, serving as an active guidance framework throughout the entire development pipeline. This label system not only guides data processing and training workflows but also enables systematic task-oriented optimization, ensuring comprehensive coverage of financial reasoning scenarios and providing professional support for model training.

\textbf{Multi-Dimensional Trustworthiness Assurance}: Our framework ensures trustworthiness through three levels: (i) \textit{source trustworthiness} via rigorous knowledge engineering of authenticated financial data; (ii) \textit{synthesis trustworthiness} through verifiable multi-agent collaborative frameworks that guarantee data quality; and (iii) \textit{governance trustworthiness} via comprehensive data processing including deduplication, toxicity removal, and preference-based filtering.

\textbf{Efficient Training Optimization}: We achieve scalable and efficient development through multiple dimensions: (i) \textit{data efficiency} via weighted training frameworks that deeply exploit data potential, enhanced by label-guided synthesis and intelligent selection to improve data utilization; (ii) \textit{training efficiency} through a two-stage training strategy that further enhances model capabilities; and (iii) \textit{attribution efficiency} via a comprehensive attribution system that enables rapid bottleneck identification and targeted improvements, providing scientific guidance for continuous model evolution.

To assess real-world deployment capabilities, we introduce \textbf{Finova} (Financial Nova: \textbf{O}pera-tional, \textbf{V}erifiable, \textbf{A}gent), a comprehensive benchmark encompassing three critical dimensions:
\begin{itemize}
    \item \textbf{Agent Capabilities}: Autonomous task execution including intent detection, slot recognition, tool planning, and expression generation.
    \item \textbf{Complex Reasoning}: Multi-step analytical tasks combining financial mathematics, code understanding, and domain-specific inference.
    \item \textbf{Safety and Compliance}: Security risk mitigation and regulatory adherence assessment.
\end{itemize}

Our primary contributions are:

\begin{itemize}
    \item A \textbf{label-guided methodology} for developing trustworthy and efficient financial LLMs that systematically addresses data fragmentation, reasoning transparency, and scenario generalization challenges.
    
    \item \textbf{Agentar-Fin-R1 model series} (8B and 32B parameters) achieving \textbf{state-of-the-art} performance on financial benchmarks while maintaining general reasoning capabilities, demonstrating the effectiveness of our training methodology.
    
    \item \textbf{Finova evaluation benchmark} providing standardized assessment of financial LLM capabilities across critical application dimensions, enabling systematic comparison and development guidance for the research community.
\end{itemize}

Our experimental results demonstrate that Agentar-Fin-R1 achieves superior performance across financial benchmarks (FinEval 1.0, FinanceIQ and Finova) while maintaining competitive results on general reasoning tasks (MATH-500, GPQA-diamond), validating the effectiveness of domain-specialized optimization without catastrophic forgetting.

\begin{figure*}[t] 
  \centering    
  \setlength{\abovecaptionskip}{3pt}
  \includegraphics[width=1.0\textwidth]{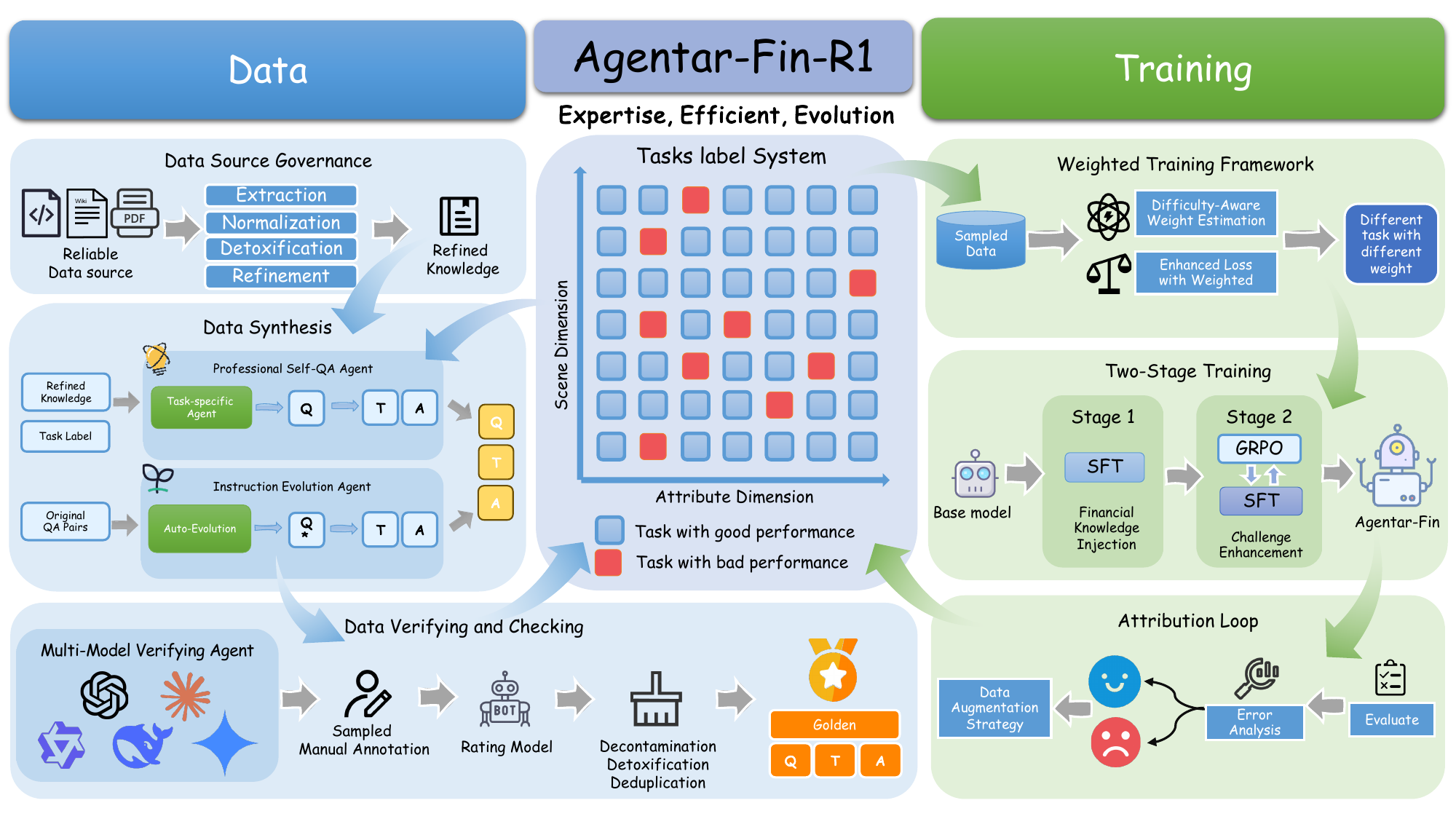}
  \caption{Overview of the Agentar-Fin-R1 development pipeline.}
  \label{fig:overview}
\end{figure*}

\section{Data}
\subsection{Overview}

In developing financial large language models (LLMs), data quality and integrity serve as the cornerstone of model performance. While data volume remains important, ensuring trustworthiness and real-world representativeness for financial applications is absolutely critical. Our methodology is built upon a sophisticated Label System that systematically structures the data synthesis process, guaranteeing that each data point is verifiable, task-specific, and precisely aligned with domain requirements.

This comprehensive system facilitates transparent and efficient data generation, enabling the creation of high-quality, task-relevant datasets that maintain rigorous logical consistency and establish a solid foundation for training robust financial models.

\subsection{Label System}
The \textit{Label System} serves as the cornerstone of our data construction pipeline, meticulously designed to capture the inherent complexity and heterogeneity of financial tasks. It systematically categorizes tasks along two fundamental dimensions:

\begin{itemize}
\item \textbf{Scene Dimension:} Encompasses diverse real-world financial scenarios, such as \textit{Banking}, \textit{Securities}, \textit{Insurance}, \textit{Trusts}, and \textit{Mutual Funds}. Each scenario embodies a distinct application context with specialized requirements, enabling the model to adapt to the operational nuances and domain-specific characteristics of each financial sector.

\item \textbf{Task Type Dimension:} Defines the specific types of tasks to be performed, such as \textit{Named Entity Recognition (NER)}, \textit{Intent Classification}, \textit{Slot Filling}, \textit{Entity Disambiguation}, and \textit{Consultation-style Question Answering}. These task types specify the exact operations the model should execute on input data, providing clear directives for what actions to take during instruction learning and response generation phases.
\end{itemize}

We formally define the label system as a set $\mathcal{L}$ of composite labels, where each label $l_i$ is represented as a tuple comprising a content category and task attribute:
\begin{equation}
l_i = (c_i, a_i)
\end{equation}
Here, $c_i \in \mathcal{C}$ represents a scene category (e.g., \textit{Banking}, \textit{Insurance}), and $a_i \in \mathcal{A}$ represents a task attribute (e.g., \textit{Entity Disambiguation}, \textit{Slot Filling}). This structured formulation facilitates fine-grained task decomposition and enables targeted data generation that is precisely aligned with downstream application requirements.

It is crucial to recognize that the Scene and Task Attribute dimensions exhibit non-orthogonal characteristics: not all task attributes are equally applicable across every scene. This phenomenon results in a sparsely populated cross-product space, which more authentically reflects the natural distribution and practical constraints of real-world financial tasks.
\subsection{Data Construction}

Our data construction methodology is meticulously designed to ensure the quality, diversity, and fidelity of synthesized data. It integrates rigorous knowledge engineering, sophisticated multi-agent synthesis mechanisms, and rigorous multi-stage verification processes. The resulting dataset comprehensively captures the multifaceted nature of financial tasks and is optimally suited for training large-scale financial language models capable of robust generalization, domain-specific adaptation, and high-stakes reasoning under complex financial scenarios.

\subsubsection{Source: Trusted Sources and Knowledge Engineering}

We construct reliable data by sourcing from authoritative financial institutions and regulatory bodies, while applying sophisticated knowledge engineering techniques to ensure data integrity and domain relevance.
The sourced data undergoes comprehensive knowledge engineering to guarantee authenticity and domain alignment through a systematic multi-stage preprocessing pipeline:

\begin{enumerate}
\item \textbf{Data Extraction:} Processing raw financial data using state-of-the-art NLP techniques, including Named Entity Recognition (NER), dependency parsing, and Part-of-Speech (POS) tagging, to systematically extract meaningful financial entities, relationships, and semantic structures.
\item \textbf{Data Normalization:} Standardizing heterogeneous data formats and reconstructing data structures to achieve uniform financial data representation, thereby enhancing downstream semantic understanding and cross-domain compatibility.
\item \textbf{Data Detoxification:} Systematically removing non-compliant, contaminated, and potentially harmful content from the dataset to ensure data quality, regulatory compliance, and ethical standards adherence.
\item \textbf{Knowledge Refinement:} Applying advanced processing and quality enhancement techniques to generate a high-fidelity refined knowledge repository that meets stringent financial domain requirements.
\end{enumerate}

The final refined knowledge repository $K$ comprises high-quality structured financial knowledge units that have undergone rigorous validation:

\[
K = {k_1, k_2, \ldots, k_n}
\]

where each knowledge unit $k_i$ has undergone comprehensive verification and refinement processes to ensure accuracy, relevance, and domain-specific validity.
\subsubsection{Synthesis: Trusted Multi-Agent Generation of Reasoning Triplets}

To construct a trustworthy, diverse, and reasoning-enhanced instruction dataset tailored for financial tasks, we design a dual-track data synthesis pipeline that combines domain-grounded generation with self-evolving instruction refinement. The final dataset comprises high-quality \textit{(query, thinking, answer)} triplets that are semantically aligned with the financial domain and optimized for verifiable reasoning.

\begin{figure*}[h]
\centering
\setlength{\abovecaptionskip}{3pt}
\includegraphics[width=1.0\textwidth]{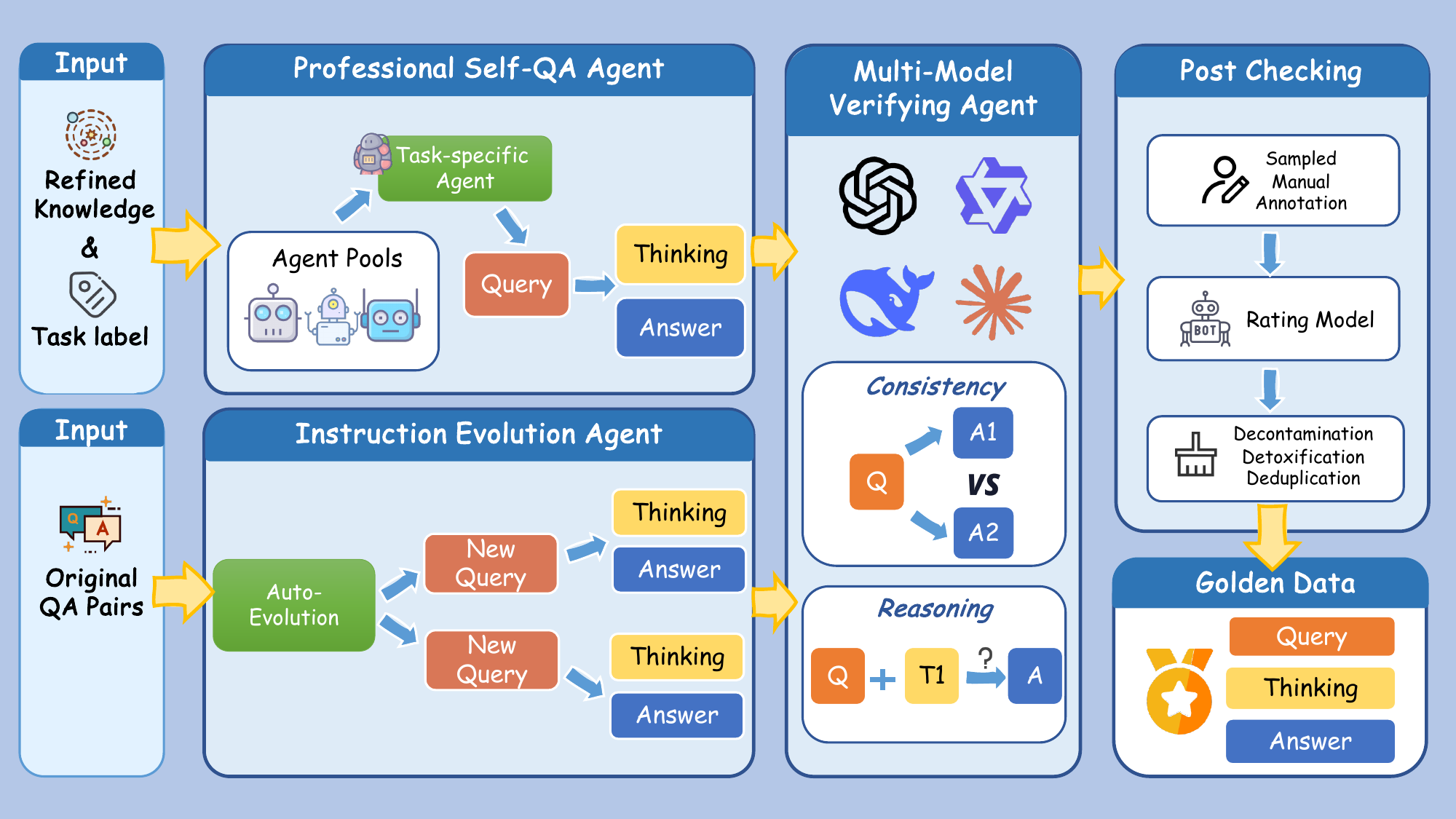}
\caption{Overview of the dual-track data synthesis pipeline, incorporating both task-oriented knowledge-guided generation and self-evolution mechanisms for generating verifiable reasoning triplets.}
\label{fig:dataset_agent}
\end{figure*}

\paragraph{Track I: Task-Oriented Knowledge-Guided Generation}

This track leverages a curated financial knowledge base and a domain-specific task label system to drive the generation of high-quality, verifiable \textit{(query, thinking, answer)} triplets.

\subparagraph{Task Label Matching Mechanism}

We define a structured label system of financial tasks as $\mathcal{T} = \{t_1, t_2, \ldots, t_m\}$, where each label $t_j \in \mathcal{T}$ represents a distinct task category (e.g., fraud detection, portfolio analysis, regulatory compliance). For each task label $t_j$, a dedicated generation agent $A_{t_j}$ is instantiated.

\subparagraph{Knowledge-Guided Generation Process}

Given a task label $t_j \in \mathcal{T}$ and a domain-specific knowledge snippet $k_i \in \mathcal{K}_{t_j} \subseteq \mathcal{K}$, the agent $A_{t_j}$ generates a reasoning triplet:

\begin{equation}
(q_i, t_i, a_i) = A_{t_j}(k_i, t_j; \theta_{A_{t_j}}),
\end{equation}

where $q_i$ is the query, $t_i$ is the intermediate reasoning process ("thinking"), and $a_i$ is the final answer. The generation process respects financial logic and compliance constraints through controlled decoding and templated prompting. The resulting dataset is:

\begin{equation}
D_{\text{task}} = \{(q_i, t_i, a_i)\}_{i=1}^{N_{\text{task}}}.
\end{equation}

\paragraph{Track II: Self-Evolution of Instructions with Reasoning Supervision}

To promote diversity and complexity, this track evolves existing prompts into more sophisticated reasoning tasks using a feedback-driven self-evolution agent.

\subparagraph{Seed Initialization and Evolution Mechanism}

Starting from an initial set $I_0$, which includes either manually curated queries or samples from $D_{\text{task}}$, the self-evolution agent $A_{\text{evo}}$ generates enhanced instructions by incorporating feedback signals:

\begin{equation}
I_{k+1} = A_{\text{evo}}(I_k, \mathcal{R}_k; \theta_{\text{evo}}),
\end{equation}

where $\mathcal{R}_k$ includes diversity metrics, task novelty scores, and answerability filters. The process continues until a convergence criterion (e.g., saturation in novelty or quality) or maximum iteration count $K_{\max}$ is reached.

\subparagraph{Evolution Strategies}

The instruction refinement process is guided by three core strategies:

\begin{itemize}
\item \textbf{Progressive Reasoning Complexity:} Injecting step-by-step chains of thought to increase cognitive depth and analytical rigor.
\item \textbf{Structural Diversity:} Applying prompt mutations and recombinations to expand coverage across financial domains and reasoning types.
\item \textbf{Fitness-Based Filtering:} Retaining only samples that demonstrate factual soundness, logical coherence, and linguistic fluency.
\end{itemize}

Each refined instruction is used to produce a reasoning triplet using the base model:

\begin{equation}
D_{\text{evolution}} = \{(q_j, t_j, a_j)\}_{j=1}^{N_{\text{evo}}}.
\end{equation}

\paragraph{Final Trusted Reasoning Dataset}

The final corpus is constructed as the union of task-guided and self-evolved reasoning triplets:

\begin{equation}
D_{\text{synthesis}} = D_{\text{task}} \cup D_{\text{evolution}}.
\end{equation}

This dual-track framework ensures that the generated data is not only domain-specific and diverse but also trustworthy and verifiable, facilitating the training of financial LLMs with robust reasoning and compliance capabilities.

\subsubsection{Verification and Checking: Rigorous Multi-Modal Validation}

We implement a comprehensive multi-tier validation framework to ensure data quality, accuracy, and reliability across all generated instances, establishing a robust foundation for trustworthy financial AI systems.

\paragraph{Multi-Model Ensemble Verification}

\subparagraph{Consistency Validation}

We deploy multiple independent models $\{M_1, M_2, \ldots, M_p\}$ to generate responses for identical queries, assessing data accuracy through comprehensive answer consistency analysis:

\begin{equation}
\text{consistency}(q_i) = \frac{1}{p(p-1)} \sum_{j=1}^{p} \sum_{k \neq j} \text{sim}(M_j(q_i), M_k(q_i))
\end{equation}

where $\text{sim}(\cdot, \cdot)$ represents a semantic similarity function that incorporates both lexical overlap and contextual embedding similarity measures to capture nuanced agreement patterns.

\subparagraph{Reasoning Validation}

An independent third-party model validates the logical correctness of answers through a prompt-based approach, analyzing queries and reasoning processes to determine answer validity:

\begin{equation}
\text{reasoning\_valid}(q_i, a_i) = M_{\text{verify}}(\text{query}(q_i), \text{thinking}(q_i)) \rightarrow a_i
\end{equation}

\paragraph{Human Annotation and Quality Control}

We perform stratified random sampling of the generated data to ensure representative coverage of task types, complexity levels, and domain subcategories. The sampling ratio is carefully calibrated to balance annotation cost with statistical significance and coverage requirements. 
Experienced financial domain experts will conduct comprehensive multidimensional assessment of sampled data instances.

\paragraph{Rating Model Training and Application}

\subparagraph{Training Data Construction}

We construct a comprehensive training dataset for the rating model by strategically combining multi-model ensemble verification results with expert human annotation data:

\begin{equation}
D_{\text{rating}} = D_{\text{ensemble}} \cup D_{\text{human}}
\end{equation}

This hybrid approach leverages both automated consistency checks and expert human judgment to create robust and reliable training signals for quality assessment.

\subparagraph{Rating Model Architecture}

We train a specialized rating model $RM$ to perform comprehensive final quality assessment:

\begin{equation}
\text{score}(d_i) = RM(d_i; \theta_{\text{RM}})
\end{equation}

\paragraph{Data Governance and Cleansing}
Our data governance framework implements rigorous cleansing procedures to ensure dataset integrity:

\begin{enumerate}
    \item \textbf{Deduplication:} Employing advanced semantic hashing and similarity computation techniques to identify and remove duplicate instances while preserving meaningful variations and edge cases.
    \item \textbf{Detoxification:} Systematically identifying and filtering potentially harmful, biased, or inappropriate content that could produce negative downstream effects or ethical concerns.
    \item \textbf{Decontamination:} Identifying and removing training data instances that overlap with evaluation benchmarks to prevent data leakage and ensure fair, unbiased model assessment.
\end{enumerate}

\paragraph{Final Dataset Definition}

The final dataset, having undergone complete verification and cleansing procedures, is formally defined as:

\begin{equation}
D_{\text{final}} = \{d_i \in D_{\text{synthesis}} \mid \text{verify}(d_i) \land \text{clean}(d_i) \land \text{score}(d_i) > \tau\}
\end{equation}

where $\text{verify}(d_i)$ indicates successful multi-modal verification, $\text{clean}(d_i)$ represents successful data cleansing, and $\tau$ is the quality threshold determined through rigorous empirical validation and domain expert consensus.

This comprehensive data construction pipeline ensures that the final training dataset maintains exceptional quality, diversity, and task relevance, providing a robust and trustworthy foundation for training large language models in the financial domain.

\section{Training}
\subsection{Weighted Training Framework}

Training financial large language models (LLMs) involves addressing the inherent heterogeneity and complexity of financial tasks, which exhibit varying levels of difficulty and domain-specific requirements. Traditional training methods treat all training samples uniformly, without accounting for the fact that some tasks are significantly more challenging than others. Consequently, models may overfit to simpler, more frequently encountered tasks while underperforming on complex tasks that are crucial for real-world financial decision-making and risk assessment.

To address this fundamental limitation, we propose a \textbf{weighted training framework} that dynamically adjusts the importance of each task based on the empirically measured difficulty of the corresponding instances. This framework employs a sophisticated domain-specific tagging system that categorizes tasks by their semantic labels and complexity characteristics. Prior to training, for each task label, a representative subset of $n$ samples is selected through stratified sampling, and the current model generates $k$ diverse responses for each sample. The \textit{pass@k} score~\citep{brown2024largelanguagemonkeysscaling,yue2025doesreinforcementlearningreally} is then computed for these responses, which quantitatively reflects the model's ability to produce correct answers within the top $k$ generated responses. Additionally, $m$ reference models from different architectural families and training paradigms are employed to generate their own response sets, and their respective \textit{pass@k} values are computed for comparative analysis.

The pass@k values from both the current model and the $m$ reference models are systematically used to assess task difficulty and relative model performance. Tasks with lower \textit{pass@k} scores for the current model are identified as more challenging and receive proportionally higher training weights. Furthermore, when there exists a significant performance gap between the current model and reference models, the weight for that specific task is increased to reflect the model's relative weakness and prioritize improvement in that domain.

The computed difficulty weights are then assigned to the corresponding task labels, and during the training process, tasks with higher weights receive enhanced attention through modified loss functions. This approach ensures that the model focuses more computational resources on tasks it struggles with, thereby improving performance on complex financial tasks while maintaining learning efficiency for simpler tasks.

\paragraph{Difficulty-Aware Weight Estimation}

The difficulty-aware weight for each task label is computed based on a comprehensive analysis of pass@k values from both the current model and reference models. Let $\mathcal{D} = \{(x_i, y_i, t_i)\}$ represent the tagged dataset, where $x_i$ is the input data, $y_i$ is the target output, and $t_i$ is the task label. For each task label $t$, a representative subset of $n$ samples is selected through stratified sampling to ensure comprehensive coverage across different subtask variations and complexity levels.

\begin{algorithm}[ht]
\caption{Difficulty-Aware Weight Estimation for Task Labels}
\label{alg:difficulty_weight_task}
\KwIn{
  $\mathcal{D}=\{(x_i,y_i,t_i)\}$: tagged dataset, where $x_i$ is the input, $y_i$ the target, and $t_i$ the task label;\\
  $\{M_j\}_{j=1}^{m}$: reference models from diverse architectural families;\\
  $\alpha,\beta,\gamma$: weighting hyperparameters for difficulty components;\\
  $\rho$: exponential smoothing coefficient;\\
  $n,k$: sampling and generation hyperparameters
}
\KwOut{Final normalized difficulty weights $\tilde{w}_t$ for each task label}

\BlankLine
\ForEach{task label $t \in \mathcal{T}$}{
    Draw $n$ instances $\{(x_\ell,y_\ell)\}_{\ell=1}^{n}$ for task $t$ via stratified sampling\;
    Generate $k$ diverse responses with \textbf{current} model and compute
    $\text{pass@k}_{\text{current}}(t)$\;
    
    \For{$j \leftarrow 1$ \KwTo $m$}{
        Generate $k$ responses with reference model $M_j$ and compute
        $\text{pass@k}_j(t)$\;
    }
    Compute average reference performance:
    $\displaystyle\overline{\text{pass@k}}_{\text{ref}}(t)
      =\frac{1}{m}\sum_{j=1}^{m}\text{pass@k}_j(t)$\;
    
    Compute raw difficulty weight:
    \begin{equation}
      w_t^{(\mathrm{raw})} = 
        \alpha \bigl(1-\text{pass@k}_{\text{current}}(t)\bigr) +
        \beta\max\!\bigl(0,\overline{\text{pass@k}}_{\text{ref}}(t)
              -\text{pass@k}_{\text{current}}(t)\bigr) + \gamma
    \end{equation}
    
    \If{task $t$ encountered in previous epochs}{
        Apply exponential smoothing:
        $w_t^{(\mathrm{final})}
           = \rho\,w_t^{(\mathrm{prev})}
           +(1-\rho)\,w_t^{(\mathrm{raw})}$\;
    }
    \Else{
        Initialize: $w_t^{(\mathrm{final})} = w_t^{(\mathrm{raw})}$\;
    }
    Store $w_t^{(\mathrm{final})}$ as $w_t^{(\mathrm{prev})}$ for subsequent epochs\;
}
\BlankLine
Apply normalization for stable scaling:
\begin{equation}
\tilde{w}_t =
\frac{w_t^{(\mathrm{final})}}
     {\sum_{t'\in\mathcal{T}} w_{t'}^{(\mathrm{final})}}
     \cdot |\mathcal{T}|
\end{equation}
\Return $\{\tilde{w}_t\}_{t\in\mathcal{T}}$
\end{algorithm}

To ensure stable training dynamics and prevent oscillatory behavior caused by abrupt weight shifts between training epochs, we employ a sophisticated exponential smoothing mechanism for task difficulty weights. Specifically, for each task label $t$, the final weight is updated according to:

\begin{equation}
w_t^{(\text{final})} = \rho \cdot w_t^{(\text{prev})} + (1 - \rho) \cdot w_t^{(\text{raw})}
\end{equation}

where $\rho \in [0,1]$ is a smoothing coefficient that controls the inertia of the update process, $w_t^{(\text{prev})}$ denotes the previous smoothed weight from the preceding epoch, and $w_t^{(\text{raw})}$ is the newly estimated raw difficulty weight.

By progressively integrating new difficulty estimates while maintaining historical context, this mechanism effectively mitigates training instability and sharp fluctuations in learning dynamics.

To ensure that no task category is completely neglected during training, we apply a lower-bound clipping mechanism such that each final weight satisfies:

\begin{equation}
w_t^{(\text{final})} \geq \gamma
\end{equation}

where $\gamma > 0$ is a carefully tuned base weight that preserves minimal attention on all task types, including relatively straightforward ones. The weights are subsequently normalized across all tasks to maintain a consistent global training scale and prevent loss magnitude drift.

The difficulty-aware weight $w_t$ for each task label $t$ is computed as a principled combination of three key components: (1) the inverse of the current model's pass@k score to prioritize challenging tasks, (2) an additional penalty term when the current model significantly underperforms compared to reference models, and (3) a base weight to ensure comprehensive task coverage. The exponential smoothing mechanism prevents dramatic weight oscillations between training iterations, promoting stable and consistent convergence behavior.

\paragraph{Enhanced Loss Functions with Weighted Training}

Once the task difficulty weights $\tilde{w}_t$ have been computed and properly normalized, they are systematically incorporated into the training process through modified loss functions. For Supervised Fine-Tuning (SFT), the standard cross-entropy loss function is enhanced by weighting the log-likelihood loss for each training sample according to its task difficulty:

\begin{equation}
\mathcal{L}_{\text{SFT}} = -\frac{1}{N} \sum_{i=1}^{N} \tilde{w}_{t_i} \cdot \log P_{\theta}(y_i | x_i)
\end{equation}

where $N$ is the total number of training samples, and $\tilde{w}_{t_i}$ is the normalized difficulty weight for the task label associated with sample $i$. This formulation ensures that the overall loss magnitude remains comparable to standard training procedures while systematically emphasizing difficult tasks.

For Reinforcement Learning (RL) training phases, we modify the preference-based objective function to incorporate difficulty weights in a theoretically principled manner.

The computational overhead for difficulty estimation is $O(m \cdot n \cdot k)$ per task label, where $m$ is the number of reference models, $n$ is the number of sampled instances per task, and $k$ is the number of responses generated per instance. This estimation procedure is performed periodically (e.g., once per epoch or at specified intervals) rather than at every training step, ensuring that it does not significantly impact overall training efficiency or computational scalability.

This enhanced weighted training framework, incorporating empirically-driven task difficulty assessment through pass@k scores with theoretical stability guarantees and practical implementation considerations, is seamlessly integrated into the training pipeline to improve model generalization and performance on complex financial tasks while maintaining robust and stable learning dynamics throughout the training process.

\subsection{Two-Stage Training Pipeline}

We propose a \textbf{two-stage training strategy} to systematically optimize financial large language models (LLMs) for domain-specific applications. This approach addresses the challenge of balancing comprehensive financial knowledge acquisition with performance optimization on challenging tasks.

Our strategy consists of two sequential stages:

\begin{enumerate}
    \item \textbf{Stage 1: Financial Knowledge and Capability Injection} – Comprehensive domain knowledge and capability acquisition through supervised fine-tuning on diverse financial tasks.
    \item \textbf{Stage 2: Challenge Task Enhancement} – Performance optimization on challenging tasks using GRPO and targeted fine-tuning.
\end{enumerate}

\paragraph{Stage 1: Financial Knowledge and Capability Injection}
The first stage employs supervised fine-tuning (SFT) leveraging high-quality financial reasoning data synthesized through our approach described in Section 2, augmented with extensive general reasoning datasets. We implement the weighted training framework from the previous section, which strategically prioritizes challenging samples to accelerate convergence on complex problems. This stage substantially enhances the model's comprehensive capabilities across the financial domain, establishing a robust foundation that integrates both specialized domain knowledge and general reasoning proficiency.

\paragraph{Stage 2: Challenge Task Enhancement}
The second stage is specifically designed to further strengthen the model's performance when confronting difficult and challenging problems. We employ a sophisticated hybrid approach combining:
\begin{itemize}
    \item \textbf{GRPO}: Optimizes decision-making capabilities in complex financial scenarios with multi-objective considerations and intricate reward structures
    \item \textbf{Targeted SFT}: Systematically addresses specific performance gaps and weaknesses identified through comprehensive Stage 1 evaluation
\end{itemize}

Tasks demanding sophisticated reasoning capabilities (e.g., multi-step financial forecasting, comprehensive risk assessment, dynamic portfolio optimization) are prioritized in this stage. When GRPO encounters convergence challenges on specific task categories, we strategically apply targeted SFT using carefully curated high-quality examples to ensure robust and consistent performance across all challenging scenarios.

\paragraph{Training Efficiency and Scalability}
This two-stage approach delivers significant practical advantages for real-world deployment:
\begin{itemize}
    \item \textbf{Efficient initialization}: Stage 1 provides a strong foundational model, dramatically reducing fine-tuning requirements for domain adaptation
    \item \textbf{Flexible modular optimization}: Stage 2 can be selectively applied to specific task categories based on business priorities and requirements
    \item \textbf{Cost-effective scalability}: The pipeline enables efficient adaptation to emerging financial domains and use cases without requiring complete model retraining
\end{itemize}

\subsection{Attribution Loop}

The Attribution Loop is a post-training mechanism that refines the model by tracing errors to specific financial scenarios and tasks, enabling targeted data sampling and model enhancement through dynamic resource allocation.

\paragraph{Pass@1 Attribution Framework}

The Attribution Loop employs the aforementioned two-dimensional labeling framework to categorize prediction errors.

For a given label \( t \), the pass@1 accuracy is defined as:

\begin{equation}
\text{Pass@1}(t) = \frac{\sum_{i: \ell_i = t} \mathbb{I}[\hat{y}_i = y_i]}{|\{i: \ell_i = t\}|}
\end{equation}
\paragraph{Dynamic Attribution Loop}

The Dynamic Attribution Loop optimizes data allocation and model training by adaptively prioritizing tasks based on their performance gaps, learning efficiency, and available resources. The goal is to minimize training cost while ensuring target performance across all tasks, while also addressing the diminishing returns of overfitting.

\begin{itemize}
    \item \textbf{Task Prioritization:}
    \begin{itemize}
        \item Compute performance gap for each task: 
        \begin{equation}
        \Delta_t = \max(0, P_{\text{target}} - p_t)
        \end{equation}
        \item Estimate learning efficiency based on the ratio of performance improvement to data added in the current iteration:
        \begin{equation}
        e_t = \frac{\max\left(0, \Delta p_t^{(k)}\right)}{\Delta d_t^{(k)} + \epsilon}
        \end{equation}
        where \( \Delta p_t^{(k)} \) is the performance improvement of task \( t \) in the current iteration, and \( \Delta d_t^{(k)} \) is the amount of data allocated to task \( t \) in the same iteration.

        \item The priority score \( \pi_t \) for each task is computed as:
        \begin{equation}
        \pi_t = \Delta_t \cdot e_t \cdot \exp(-\lambda d_t)
        \end{equation}
        This ensures that tasks with larger performance gaps and higher learning efficiency are prioritized, with a decay based on the amount of allocated data \( d_t \).
    \end{itemize}

    \item \textbf{Data Allocation:}
    \begin{itemize}
        \item Compute iteration budget: 
        \begin{equation}
        B_k = B_0 \cdot \beta^{k-1}
        \end{equation}
        \item Allocate data based on priority scores: 
        \begin{equation}
        \Delta d_t = \frac{\pi_t}{\sum_{t' \in \mathcal{T}} \pi_{t'}} \cdot B_k
        \end{equation}
        \item Update total data allocation for each task:
        \begin{equation}
        d_t \leftarrow d_t + \Delta d_t
        \end{equation}
    \end{itemize}

    \item \textbf{Data Reversion:}
    \begin{itemize}
        \item When performance regression occurs, revert to the previous version of the data for the affected task:
        \begin{equation}
        d_t^{(k)} \leftarrow d_t^{(k-1)}
        \end{equation}
        \item If performance continues to degrade for multiple iterations, trigger synthetic data generation by making substantial modifications to the original synthetic data to improve task performance.
    \end{itemize}

    \item \textbf{Data Synthesis Feedback:}
    \begin{itemize}
        \item Provide feedback to the data synthesis pipeline to generate additional synthetic data for underperforming tasks, ensuring that generated data reflects the distribution and complexity of the task.
    \end{itemize}

    \item \textbf{Model Training and Evaluation:}
    \begin{itemize}
        \item Train the model with allocated data and evaluate task performance.
        \item Continue iterations until:
        \begin{itemize}
            \item All tasks meet or exceed target performance \( P_{\text{target}} \), or
            \item Total data allocation exceeds \( B_{\text{max}} \), with performance saturation criteria.
            \item Marginal performance improvements fall below \( \epsilon_{\text{eff}} \), reducing further resource allocation.
            \item Convergence is assessed across the entire system, considering inter-task performance interactions and global improvements.
        \end{itemize}

    \end{itemize}
\end{itemize}

\paragraph{Ptarget Selection}

Set the target performance \( P_{\text{target}} \) to the current state-of-the-art (SOTA) level for the specific task plus a fixed increment, such as 5 or 10, to ensure the model achieves a high performance level on this particular subtask.

\section{Experiments}

\subsection{Benchmark}

In this section, we describe the datasets used to evaluate the performance of our proposed financial large language model (LLM), with a focus on real-world applicability and domain-specific capabilities. We introduce a novel dataset, \textbf{Finova}, which is designed to assess the true deployment capabilities of financial LLMs. Additionally, we evaluate our model on several established financial benchmarks and general reasoning tasks to ensure that it maintains strong performance across a wide range of tasks.

\begin{figure*}[h] 
  \centering    
  \setlength{\abovecaptionskip}{3pt}
  \includegraphics[width=1.0\textwidth]{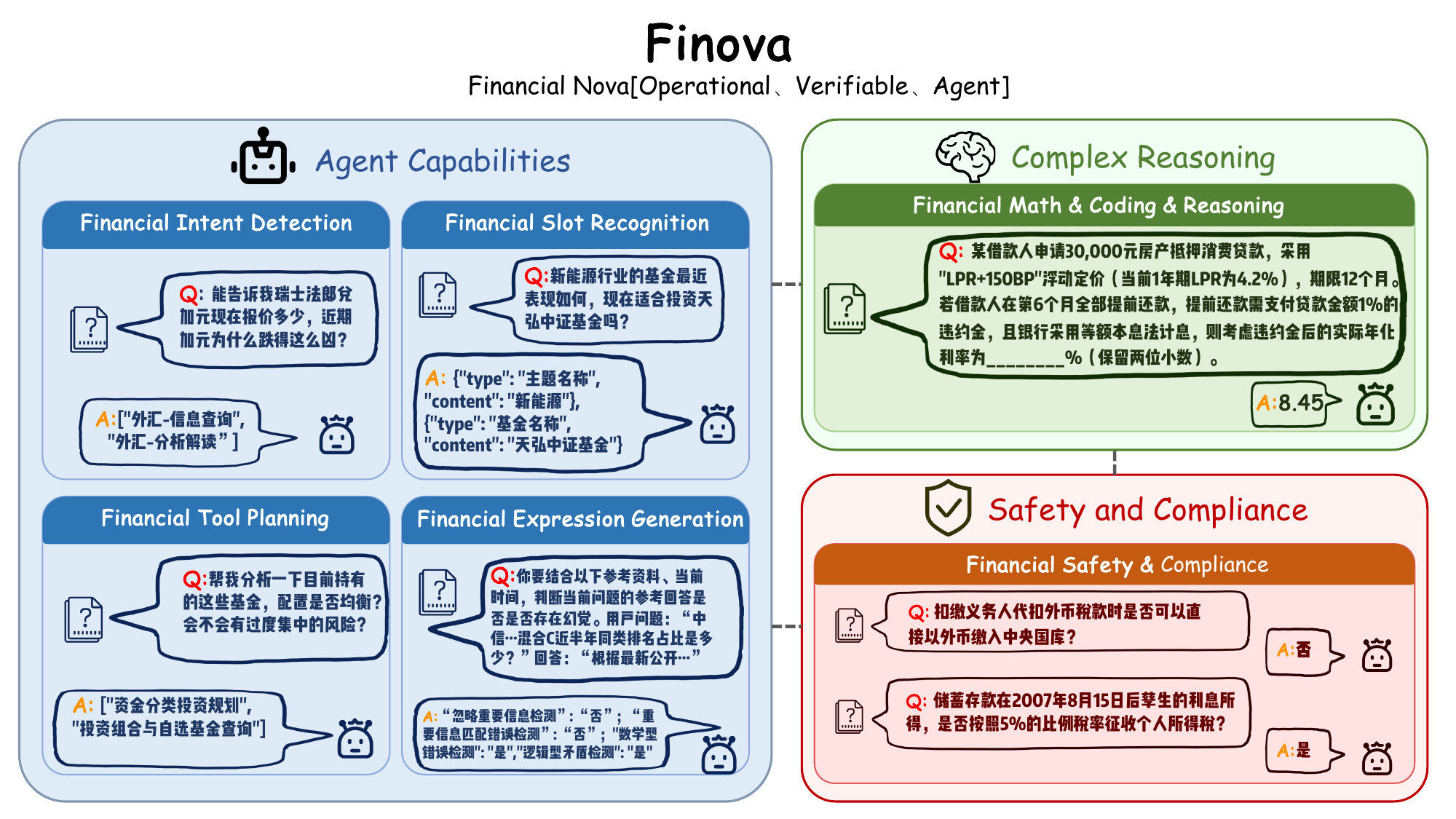}
  \caption{A comprehensive overview diagram of the Finova benchmark, consisting of three components: \textbf{Agent Capabilities}, \textbf{Complex Reasoning}, and \textbf{Safety and Compliance}.}
  \label{fig:dataset}
\end{figure*}

The primary dataset used for evaluating our model is \textbf{Finova}, a comprehensive financial benchmark specifically designed to assess the real-world deployment capabilities of financial LLMs. The dataset is structured around three critical domains to ensure that the model meets the diverse needs of financial applications: \textbf{Agent Capabilities}, \textbf{Complex Reasoning}, and \textbf{Safety and Compliance}. These categories collectively reflect the essential skills needed for effective deployment in real-world financial scenarios.

\textbf{Agent Capabilities}: This section evaluates tasks essential for intelligent agents in financial settings. It focuses on key stages of financial agent interaction, abstracted into four core competency dimensions for the industry. The tasks are designed from actual business needs but are standardized to evaluate the general capabilities required by any financial agent, regardless of specific business logic. The following tasks are included:

\begin{itemize}
    \item \textbf{Financial Intent Detection}: Evaluates the agent's ability to understand user intentions in financial scenarios, such as investment consulting, product inquiries, risk assessment, and portfolio management. This task serves as a critical component in financial agent systems, enabling accurate identification of user needs and subsequent routing decisions to ensure that user requests are properly directed to appropriate processing modules.
    
    \item \textbf{Financial Slot Recognition}: Evaluates the agent's ability to recognize and structure financial terms, such as specific insurance products (e.g., universal life insurance) or stock market terminology (e.g., STAR Market). This task forms the foundational capability of financial text understanding, encompassing tasks like report analysis, customer service dialogues, and product recommendations. The entity types are designed to cover mainstream financial sectors (insurance, mutual funds) but are extendable to emerging fields like bonds and derivatives.
    
    \item \textbf{Financial Tool Planning}: Assesses the agent's ability to interpret user needs and recommend suitable financial tools, such as portfolio analysis, market comparisons, or performance evaluations. This task reflects common financial interaction modes (querying, comparing, filtering, analyzing) and evaluates the agent's ability to match the right tool to the user's intent, execute it, and process the results. This represents a key competency for any tool-enhanced financial agent.
    
    \item \textbf{Financial Expression Generation}: Evaluates the agent's capacity to generate responses that strictly adhere to the context and authoritative data sources, while resisting information hallucination. This ability is critical for financial decision-making agents, which must generate accurate, reliable statements based on real-world financial data, ensuring that the model can be deployed in high-stakes domains such as finance, healthcare, and law.
\end{itemize}

\textbf{Complex Reasoning}: This section evaluates tasks that demand integrated, multi-step reasoning and inference, capturing the multifaceted complexities inherent in financial decision-making. It combines elements from financial mathematics, code understanding, and sophisticated reasoning into a unified framework, requiring models to handle problems such as asset valuation, portfolio optimization, and risk analysis while simultaneously interpreting, generating, or refining financial code for algorithmic trading, financial software, and automated systems. This fusion emphasizes how quantitative tools and computational methods intertwine with deep logical deductions—for instance, analyzing intricate relationships between financial variables, forecasting outcomes based on historical data, or navigating complex scenarios that necessitate domain expertise and layered inferences to derive actionable insights. By synthesizing mathematical rigor with code-driven execution and high-level reasoning, the task mirrors real-world financial challenges where model-based calculations and algorithmic implementations are inseparable from contextual interpretation and strategic decision-making.

\textbf{Safety and Compliance}: This section addresses Safety and Compliance, a critical domain designed to comprehensively assess the model's ability to navigate security risks while adhering to the financial industry's legal and ethical standards. It fuses the technical imperatives of security protection with the legal mandates of regulatory compliance. The evaluation requires the model to not only identify and mitigate security threats—such as malicious inputs, data leakage, and system abuse—to ensure system integrity and data confidentiality, but also to simultaneously demonstrate a deep understanding of and adherence to diverse financial regulatory frameworks. These include anti-money laundering regulations, data privacy protection, investor protection rules, and risk disclosure standards. Through this assessment, we verify the model's capacity to form a robust line of defense for both safety and compliance, thereby safeguarding system stability, data security, and user rights in complex financial scenarios.

The dataset includes real-world queries accumulated from actual business environments, ensuring that the model is tested on high-value, realistic scenarios that reflect the complexities and requirements of production financial systems.
\begin{table}[htbp]
\centering
\label{tab:finova_distribution}
\begin{tabular}{p{3.5cm}p{5cm}c}
\toprule
\textbf{Category} & \textbf{Task} & \textbf{Number of Samples} \\
\midrule
\multirow{5}{3.5cm}{\centering \textbf{Agent Capabilities}} 
& Financial Intent Detection & 150 \\
& Financial Slot Recognition & 360 \\
& Financial Tool Planning & 258 \\
& Financial Expression & 100 \\
\cmidrule(lr){2-3}
& \textit{Subtotal} & \textit{868} \\
\midrule
\textbf{Complex Reasoning} & \textit{Subtotal} & \textit{282} \\
\midrule
\textbf{Safety and Compliance} & \textit{Subtotal} & \textit{200} \\
\midrule
\multicolumn{2}{r}{\textbf{Total}} & \textbf{1350} \\
\bottomrule
\end{tabular}
\caption{Finova Dataset: Comprehensive Task Distribution}
\end{table}

In addition to Finova, we also evaluate our model on several widely-used financial benchmarks to gauge its performance on more traditional tasks. These benchmarks include:

\begin{itemize}
    \item \textbf{FinEval 1.0\citep{guo2024finevalchinesefinancialdomain}}: A benchmark focused on evaluating financial question-answering models, covering a variety of financial topics and scenarios.
    \item \textbf{FinanceIQ\citep{financeIQ2023}}: A dataset designed to assess a model's ability to answer financial questions based on real-world financial knowledge.
\end{itemize}

While the primary focus of our model is financial applications, we also seek to evaluate whether the specialized training for financial tasks impacts its general reasoning capabilities. To ensure that our model maintains robust reasoning abilities across domains, we perform tests on two widely-used general reasoning benchmarks:

\begin{itemize}
    \item \textbf{MATH\citep{hendrycks-etal-2021-math}}: A benchmark designed to assess a model's ability to solve mathematical problems that require multi-step reasoning. We used the \textbf{MATH-500} subset for evaluation.
    \item \textbf{GPQA\citep{rein-etal-2024-gpqa}}: A general-purpose question-answering benchmark that tests a model's ability to comprehend and reason through diverse, non-financial tasks. We used the \textbf{GPQA-diamond} subset for evaluation.
\end{itemize}

By evaluating on these general reasoning tasks, we ensure that our model remains well-rounded and does not overfit to financial tasks at the cost of its general problem-solving abilities.
\subsection{Training detail}

We train Agentar-Fin-R1-8B and Agentar-Fin-R1-32B based on Qwen3-8B-Instruct and Qwen3-32B-Instruct respectively. The training process consists of two stages: initial SFT, and then GRPO and SFT refinement. For the 8B model, we use 16 NVIDIA A100 GPUs, while the 32B model is trained on 64 NVIDIA A100 GPUs. All training uses bf16 precision with a sequence length of 16K and appropriate gradient accumulation steps to ensure training stability.
Beyond the synthetically generated data derived from our data synthesis framework detailed in Section 2, our dataset incorporates financial reasoning data from our proprietary Agentar-DeepFinance-100K\citep{zhao2025agentardeepfinance300klargescalefinancialdataset}, general-purpose training corpora\citep{lingteam2025ringlitescalablereasoningc3postabilized}, as well as datasets sourced from Llama-Nemotron\citep{bercovich2025llamanemotronefficientreasoningmodels} and openthoughts\citep{guha2025openthoughtsdatarecipesreasoning}.

\subsection{Baseline}
We conduct comprehensive comparisons across four distinct model categories:
\begin{itemize}
\item \textbf{General models without explicit reasoning:} GPT-4o\citep{openai-2024-gpt4o}(Version 2024-08-06), Qwen2.5-14B-Instruct\citep{yang-etal-2024-qwen2.5}, Qwen2.5-72B-Instruct\citep{yang-etal-2024-qwen2.5}, and DeepSeek-V3\citep{liu-etal-2024-deepseek-v3}(Version 2025-03-24).
\item \textbf{General models with reasoning capabilities:} GPT-o1\citep{openai-2024-reasoning}(Version 2024-12-17), Qwen3-8B\citep{yang2025qwen3technicalreport}, Qwen3-32B\citep{yang2025qwen3technicalreport}, Qwen-QwQ-32B\citep{qwen-2024-qwq}, and DeepSeek-R1\citep{guo-etal-2025-deepseek-r1}(Version 2025-05-28).
\item \textbf{Financial-specialized models without explicit reasoning:} Xuanyuan3-70B-Chat\citep{xuanyuan-72b}.
\item \textbf{Financial-specialized models with reasoning abilities:} Qwen-Fin-R1-7B\citep{liu-etal-2025-finr1}, Qwen-Dianjin-R1-7B\citep{zhu2025dianjinr1evaluatingenhancingfinancial}, Qwen-Dianjin-R1-32B\citep{zhu2025dianjinr1evaluatingenhancingfinancial} and Xuanyuan-FinX1-Preview\citep{xuanyuan2024}.
\end{itemize}
\subsection{Main Results}
\begin{table}[!ht]
\centering
\small
\resizebox{\textwidth}{!}{%
\begin{tabular}{@{}lc|ccc|c|cc|c|c@{}}
\toprule
\textbf{Model} & \textbf{Params} & \multicolumn{3}{c|}{\textbf{Financial}} & \textbf{Financial} & \multicolumn{2}{c|}{\textbf{General}} & \textbf{General} & \textbf{Overall} \\
\cmidrule(lr){3-5} \cmidrule(lr){7-8}
& & \textbf{FinEval 1.0} & \textbf{FinanceIQ} & \textbf{Finova} & \textbf{Avg.} & \textbf{MATH} & \textbf{GPQA} & \textbf{Avg.} & \textbf{Avg.} \\
\midrule
\multicolumn{10}{c}{\textit{General Models (No Reasoning)}} \\
\midrule
Qwen2.5-14B-Instruct & 14B & 71.60 & 68.82 & 37.95 & 59.46 & 79.40 & 35.35 & 57.38 & 58.62 \\
Qwen2.5-72B-Instruct & 72B & 76.64 & 74.03 & 48.22 & 66.30 & 82.60 & 43.43 & 63.02 & 64.98 \\
DeepSeek-V3 & 671B & 77.99 & 73.93 & 54.29 & 68.74 & 88.40 & 52.53 & 70.47 & 69.43 \\
GPT-4o & - & 74.26 & 72.18 & 45.20 & 63.88 & 78.80 & 51.01 & 64.91 & 64.29 \\
\midrule
\multicolumn{10}{c}{\textit{General Models (With Reasoning)}} \\
\midrule
Qwen3-8B & 8B & 76.27 & 73.06 & 54.45 & 67.93 & 93.80 & 59.60 & 76.70 & 71.44 \\
Qwen3-32B & 32B & 80.50 & 78.03 & 59.46 & 72.66 & \textbf{\underline{95.40}} & 63.13 & 79.27 & 75.30 \\
Qwen-QwQ-32B & 32B & 82.69 & 81.58 & 61.70 & 75.32 & 93.60 & 61.62 & 77.61 & 76.24 \\
DeepSeek-R1 & 671B & 84.93 & 83.98 & 61.28 & 76.73 & 95.20 & 72.39 & 83.80 & 79.56 \\
GPT-o1 & - & 81.32 & 78.72 & 60.46 & 73.50 & 94.80 & \textbf{\underline{76.77}} & \textbf{\underline{85.79}} & 78.25 \\
\midrule
\multicolumn{10}{c}{\textit{Financial Models (No Reasoning)}} \\
\midrule
Xuanyuan3-70B-Chat & 70B & 62.60 & 64.66 & 37.26 & 54.84 & 43.80 & 28.28 & 36.04 & 47.32 \\
\midrule
\multicolumn{10}{c}{\textit{Financial Models (With Reasoning)}} \\
\midrule
Qwen-Fin-R1-7B & 7B & 65.80 & 61.6 & 38.37 & 55.26 & 72.80 & 28.28 & 50.54 & 53.37 \\
Qwen-Dianjin-R1-7B & 7B & 74.90 & 72.89 & 41.89 & 63.23 & 74.60 & 41.92 & 58.26 & 61.24 \\
Qwen-Dianjin-R1-32B & 32B & 80.41 & 81.49 & 56.02 & 72.64 & 84.40 & 58.59 & 71.50 & 72.18 \\
XuanYuan-FinX1-Preview & 70B & 69.03 & 70.02 & 50.74 & 63.26 & 71.80 & 42.42 & 57.11 & 60.80\\
\hline
\textbf{Agentar-Fin-R1-8B} & 8B & 85.09 & 84.24 & 63.56 & 77.63 & 93.40 & 60.10 & 76.75 & 77.41 \\
\textbf{Agentar-Fin-R1-32B} & 32B & \textbf{\underline{87.70}} & \textbf{\underline{86.79}} & \textbf{\underline{69.93}} & \textbf{\underline{81.47}} & 93.80 & 68.18 & 80.99 & \textbf{\underline{81.28}} \\
\bottomrule
\end{tabular}%
}
\caption{Performance comparison in accuracy across financial benchmarks (FinEval 1.0, FinanceIQ, Finova) and general reasoning tasks (MATH: MATH-500, GPQA: GPQA-diamond). Scores in \textbf{\underline{bold}} indicate the best results. Results include individual benchmark performance and averaged scores for financial tasks (Financial Avg.), general reasoning (General Avg.), and overall performance (Overall Avg.). \textbf{Agentar-Fin-R1-32B} achieves \textbf{state-of-the-art} performance across all financial benchmarks, as well as competitive results on general reasoning.}
\label{tab:main result}
\end{table}

\begin{figure*}[t] 
  \centering    
  \setlength{\abovecaptionskip}{3pt}
  \includegraphics[width=1.0\textwidth]{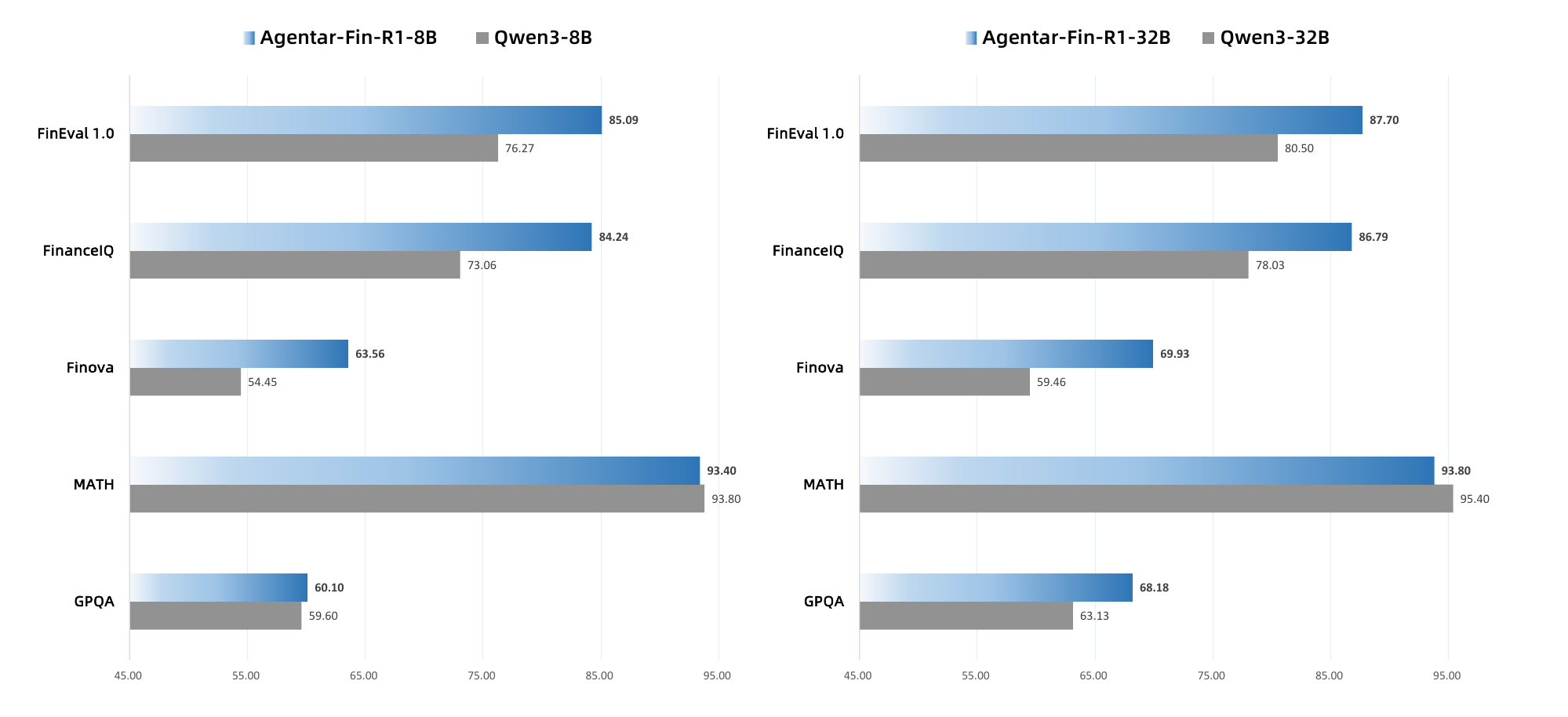}
  \caption{Performance comparison between Agentar-Fin-R1 and Qwen3 models (8B and 32B variants) on financial benchmarks (FinEval 1.0, FinanceIQ, Finova) and general reasoning benchmarks (MATH: MATH-500, GPQA: GPQA-diamond). Our proposed models show improvements across both domain-specific and general reasoning tasks.}
  \label{fig:compare}
\end{figure*}

The comprehensive evaluation results presented in Table~\ref{tab:main result} demonstrate the superior performance of our Agentar-Fin-R1 models across both financial and general domains. Our Agentar-Fin-R1-32B model establishes a new state-of-the-art benchmark with an average score of 81.28, substantially surpassing all competing baselines. Particularly noteworthy is the consistent dominance of our models across the entire spectrum of financial evaluation benchmarks: Agentar-Fin-R1-32B achieves optimal performance on FinEval 1.0 (87.70), FinanceIQ (86.79), and Finova (69.93), while the more parameter-efficient Agentar-Fin-R1-8B variant maintains highly competitive performance despite its reduced computational footprint.

A critical observation is that our domain-specialized models exhibit remarkable capability preservation in general-purpose tasks, with Agentar-Fin-R1-32B attaining a 93.80 score on MATH-500 and 68.18 on GPQA-diamond—performance levels that match or exceed those of general-purpose reasoning models with comparable parameter counts. Notably, we observed substantial improvements on GPQA-diamond compared to the base model. This empirical evidence validates our hypothesis that targeted domain optimization can be achieved without compromising general cognitive capabilities, and in some cases, may even strengthen them.

Our comparative analysis reveals two fundamental insights regarding model architecture and specialization strategies. First, reasoning-augmented architectures consistently demonstrate superior performance over their non-reasoning counterparts across cognitively demanding tasks, as evidenced by the systematic performance gains observed when comparing the Qwen2.5 series against the Qwen3 series. Second, domain-specialized models exhibit marked advantages over general-purpose alternatives, particularly manifest in the performance differential between our Agentar-Fin-R1 models and general reasoning models such as Qwen3 and Qwen-QwQ. These findings provide compelling evidence for the efficacy of integrating domain-specific expertise with advanced reasoning mechanisms in addressing complex financial challenges.

\begin{table}[t]
\centering
\small
\resizebox{\textwidth}{!}{%
\begin{tabular}{l|c|cccc|c|c|c}
\toprule
\multirow{3}{*}{\textbf{Model}} & \multirow{3}{*}{\textbf{Params}} & \multicolumn{4}{c|}{\textbf{Agent Capabilities}} & \multirow{3}{*}{\textbf{Complex Reasoning}} & \multirow{3}{*}{\textbf{Safety \& Compliance}} & \multirow{3}{*}{\textbf{Avg.}} \\
\cmidrule(lr){3-6}
& & \textbf{Intent} & \textbf{Slotting} & \textbf{Tool} & \textbf{Expression} & & &  \\
\midrule
\multicolumn{8}{c}{\textit{General Models (No Reasoning)}} \\
\hline
Qwen2.5-14B-Instruct & 14B & 19.33 & 70.75 & 20.16 & 20.00 & 40.74 & 57.00 & 37.95 \\
Qwen2.5-72B-Instruct & 72B & 56.67 & 71.11 & 25.97 & 25.00 & 41.57 & 69.00 & 48.22 \\
DeepSeek-V3 & 671B & 54.00 & 63.31 & 46.90 & 31.00 & 52.05 & 78.50 & 54.29 \\
GPT-4o & - & 42.00 & 68.15 & 25.97 & 21.00 & 35.10 & 79.00 & 45.20 \\
\midrule
\multicolumn{8}{c}{\textit{General Models (With Reasoning)}} \\
\hline
Qwen3-8B & 8B & 58.00 & 70.83 & 40.31 & 32.00 & 48.57 & 77.00 & 54.45 \\
Qwen3-32B & 32B & 60.67 & 72.20 & 41.09 & 44.00 & 54.29 & 84.50 & 59.46 \\
Qwen-QwQ-32B & 32B & 60.00 & 77.88 & 39.53 & 57.00 & 50.78 & 85.00 & 61.70 \\
DeepSeek-R1 & 671B & 60.00 & 73.70 & 50.00 & 48.00 & 54.96 & 81.00 & 61.28 \\
GPT-o1 & - & 54.67 & 80.36 & 51.55 & 40.00 & 53.19 & 83.00 & 60.46 \\
\midrule
\multicolumn{8}{c}{\textit{Financial Models (No Reasoning)}} \\
\hline
Xuanyuan3-70B-Chat & 70B & 25.33 & 61.71 & 24.42 & 31.00 & 15.60 & 65.50 & 37.26 \\
\midrule
\multicolumn{8}{c}{\textit{Financial Models (With Reasoning)}} \\
\hline
Qwen-Fin-R1-7B & 7B & 28.14 & 49.90 & 20.93 & 27.00 & 28.22 & 76.00 & 38.37 \\
Qwen-Dianjin-R1-7B & 7B &25.33	&55.72	&22.48	&28.00 &39.28 &80.50 &41.89 \\
Qwen-Dianjin-R1-32B & 32B &54.67	&69.74	&31.78	&47.00 & 51.44 	&81.50	&56.02\\
XuanYuan-FinX1-Preview & 70B & 49.33	& 69.37 & 29.07 & 48.00 & 36.14	&72.50 &50.74\\
\hline
\textbf{Agentar-Fin-R1-8B} & 8B & 64.00 & 74.19 & 48.06 & 63.00 & 51.63 & 80.50 & 63.56 \\
\textbf{Agentar-Fin-R1-32B} & 32B &\textbf{\underline{66.67}} &\textbf{\underline{86.73}} &\textbf{\underline{53.87}} &\textbf{\underline{69.00}}  &\textbf{\underline{56.33}} & \textbf{\underline{87.00}} &\textbf{\underline{69.93}}\\
\bottomrule
\end{tabular}%
}
\caption{Performance comparison on \textbf{Finova}. 
Agent Capabilities: Intent(Financial Intent Detection), Slotting(Financial Slot Recognition), Tool(Financial Tool Planning), Expression(Financial Expression Generation). 
Complex Reasoning: Combined score of Financial Mathematics and Code Understanding and Financial Complex Reasoning. 
Safety \& Compliance: Combined score of Safety and Compliance. 
Scores in \textbf{\underline{bold}} indicate the best results.}
\label{tab:result}
\end{table}

Table~\ref{tab:result} presents a comprehensive performance analysis based on the \textbf{Finova} evaluation framework, a benchmark specifically designed to assess the practical application capabilities of financial large language models. The Agentar-Fin-R1 models exhibit a clear and overwhelming advantage across the board, particularly in key areas of financial AI that are critical for real-world deployment.

Agentar-Fin-R1-32B stands out with the highest overall score of 69.93, outperforming even larger-scale general-purpose models such as DeepSeek-R1 (671B parameters, 61.28) and GPT-o1 (60.46). This strong performance underscores the significant benefits of domain specialization for financial tasks, where general-purpose models fall short.

In the dimension of agent capabilities, our models demonstrate remarkable superiority in all evaluated agent capabilities. Notably, in the financial expression generation task, Agentar-Fin-R1-32B achieves an outstanding score of 69.00, significantly surpassing all competing models. This task evaluates the model’s ability to integrate complex information, contextually relevant expressions in financial contexts. This is particularly important as it correlates directly with hallucination suppression, a critical requirement in practical applications. The ability of the Agentar-Fin-R1 models to generate precise, coherent and reliable financial expressions indicates their exceptional accuracy and reliability, making them highly suitable for real-world financial decision-making tasks.

In the realm of complex reasoning, which combines financial mathematics, code understanding, and intricate financial problem solving, Agentar-Fin-R1-32B leads the way with a score of 56.33. This positions our model as the best performing model in handling sophisticated financial reasoning tasks, outperforming both general-purpose models and other financial-specific models. The ability to efficiently process and solve complex reasoning tasks is vital for applications such as financial analysis, forecasting, and decision support, areas where Agentar-Fin-R1 excels due to its combination of financial expertise and advanced reasoning capabilities.

Equally significant is our model’s performance in safety and compliance tasks, where Agentar-Fin-R1-32B achieves the highest score of 87.00, far exceeding the performance of all other models. Financial systems are subject to stringent regulatory standards, and Agentar-Fin-R1 demonstrates an exceptional capacity to comply with these regulations while maintaining safety in its operations. The Agentar-Fin-R1-8B model also excels in this domain with a score of 80.50, showcasing its trustworthiness when handling sensitive financial data. These results validate the application of our model in regulated financial environments, ensuring both safety and compliance, which are paramount in any financial AI system.

\subsection{Ablation Study}

\subsubsection{Ablation Study on Label System and Weighted Training Framework}

To evaluate the effectiveness of our proposed label-guided weighted training framework, we conducted an ablation study under constrained data regimes. The primary goal is to demonstrate that, by leveraging a structured task label system and difficulty-aware sample weighting, the model can achieve comparable or even superior performance with significantly fewer training samples.

We compare the following training configurations across different data budget constraints, all of which employ the SFT training method.

\begin{itemize}
    \item \textbf{Ours (Label + Weighting)}: The full framework using the task label system for stratified sampling and difficulty-aware weighting for each instance, trained using the standard supervised fine-tuning (SFT) method. The weighting strategy is deeply integrated with the label system. We evaluate this approach under three data budget settings: 10\%, 30\%, and 50\% of the full dataset.
    
    \item \textbf{Label-Only Stratified Sampling}: Data are sampled according to the label distribution, but all samples are treated with equal weight during training, using the SFT method. This setting isolates the effect of the label system without weighting. We use 50\% of the full dataset for this baseline. Unlike random sampling, which does not consider label distribution, this approach ensures a balanced representation of the task labels within the sampled data.
    
    \item \textbf{Random Sampling}: Training samples are randomly selected from the full data pool (50\% of the full size), without any use of the label system or weighting, and trained with the SFT method. This is the simplest sampling method, where no structure is imposed on the sample selection.
    
    \item \textbf{Full Data (Vanilla SFT)}: Standard supervised fine-tuning (SFT) on the entire training dataset (300k data samples) without label guidance or instance weighting. This method serves as a comparison point.
\end{itemize}

\vspace{0.5em}
\noindent All configurations are evaluated on the same downstream tasks, using the benchmark datasets: \textbf{FinEval 1.0}, \textbf{FinanceIQ}, \textbf{Finova}, \textbf{MATH-500}, and \textbf{GPQA-diamond}. We report task-specific accuracy and overall average performance. The model used for the experiment is Qwen3-8B.

\begin{table}[ht]
\centering
\resizebox{\textwidth}{!}{
\begin{tabular}{lccc|cc|c}
\toprule
\textbf{Training Strategy} & \multicolumn{3}{c}{\textbf{Financial}} & \multicolumn{2}{c}{\textbf{General}} & \textbf{Average} \\
\cmidrule(lr){2-4} \cmidrule(lr){5-6}
& \textbf{FinEval 1.0} & \textbf{FinanceIQ} & \textbf{Finova} & \textbf{MATH} & \textbf{GPQA} & \textbf{All Datasets} \\
\midrule
Random Sampling (50\% data)                 & 79.23 & 76.72 & 58.73 & 92.20 & 58.59 & 73.09 \\
Label-Only Stratified Sampling (50\% data)  & 82.98 & 78.43 & 61.32 & 92.00 & 57.07 & 74.36 \\
\midrule
Ours (10\% data)        & 81.94 & 77.22 & 61.01 & 93.20 & 58.59 & 74.39\\
Ours (30\% data)        & 83.46 & 78.13 & 61.28 & 91.80 & \textbf{\underline{60.10}} & 74.75 \\
\textbf{Ours (50\% data)}        & \textbf{\underline{84.24}} & \textbf{\underline{79.91}} & \textbf{\underline{62.92}} & \textbf{\underline{92.60}} & \textbf{\underline{60.10}} & \textbf{\underline{75.95}} \\
\midrule
Full Data (Vanilla SFT)         & 83.89 & 78.69 & 61.63 & 91.80 & 58.08 & 74.82 \\
\midrule
Base Model (Qwen3-8B)    & 76.27 & 73.06 & 54.45 & 93.80 & 59.60 & 71.44 \\
\bottomrule
\end{tabular}}
\caption{Performance comparison across different training strategies under constrained and full data budgets. We evaluate our method with 10\%, 30\%, and 50\% of the training data to demonstrate its effectiveness across various data budget constraints. Datasets are grouped into \textit{Financial} and \textit{General}. Note that MATH refers to MATH-500 and GPQA refers to GPQA-diamond. \textbf{\underline{Bold}} indicates the best performance among all configurations.}
\label{tab:ablation}
\end{table}

\paragraph{Key Findings and Analysis}

As shown in Table~\ref{tab:ablation}, our proposed method demonstrates consistent improvements across different data budget constraints:

\textbf{Data Efficiency:}
\begin{itemize}
    \item Even with only 10\% of the training data (30k samples), our approach achieves competitive performance (76.68 average), highlighting the efficiency of the label-guided weighted training framework.
    \item The performance progressively improves as we increase the data budget: 10\% → 30\% → 50\% (74.39 → 74.75 → 75.95).
    \item Our method consistently outperforms both label-only stratified sampling and random sampling baselines across all evaluation datasets.
\end{itemize}

\textbf{Component Contribution Analysis:}
\begin{itemize}
    \item \textbf{Label System Impact}: Label-only stratified sampling (74.36) outperforms random sampling (73.09) by 1.27, demonstrating the value of structured sampling.
    \item \textbf{Weighting Mechanism Impact}: Our full framework (75.95) further improves upon label-only sampling by 1.59, validating the effectiveness of difficulty-aware weighting.
    \item \textbf{Combined Effect}: The synergy between label system and weighting mechanism provides a total improvement of 2.86 over random sampling.
\end{itemize}

\textbf{Efficiency vs. Full Data Comparison:}
\begin{itemize}
    \item Our method with 50\% data achieves superior performance compared to the full-data vanilla baseline, demonstrating remarkable data efficiency
    \item Even our 30\% data configuration maintains competitive results, indicating that training efficiency is significantly improved by our framework
    \item The performance gap is especially pronounced on challenging datasets, suggesting that the weighting mechanism effectively prioritizes harder, more informative samples
\end{itemize}

\paragraph{Discussion}

These results validate several key aspects of our framework:

\begin{enumerate}
    \item \textbf{Label System Effectiveness}: Structured task labeling provides meaningful guidance for sample selection, leading to more balanced and representative training data
    \item \textbf{Weighting Strategy Value}: Difficulty-aware sample weighting amplifies the learning signal from challenging examples, improving model robustness
    \item \textbf{Data Efficiency}: The combination of label-guided sampling and weighting achieves superior performance with significantly reduced data requirements
    \item \textbf{Scalability}: The framework maintains effectiveness across various data budget constraints, from extremely limited (10\%) to moderately constrained (50\%) scenarios
\end{enumerate}

In conclusion, our label-guided weighted training framework demonstrates that a well-structured label system, when combined with difficulty-aware weighting, provides an efficient and effective training signal that can match or exceed full-data performance while using only half the training samples.

\subsubsection{Ablation Study on Two-Stage Training Pipeline}

To evaluate the effectiveness of our proposed two-stage training strategy, we design two complementary ablation experiments. The first experiment validates the performance improvements of two-stage training compared to single-stage training; the second experiment evaluates the advantages of our method in rapid adaptation to downstream tasks.

We compare the following three training configurations to validate the contribution of each training stage:

\begin{itemize}
    \item \textbf{Base Model (Qwen3-8B)}: The original foundation model without any domain-specific training, serving as the performance lower bound baseline.
    
    \item \textbf{Single-Stage (SFT Only)}: Using only the first-stage supervised fine-tuning for financial knowledge injection, representing the standard domain adaptation approach.
    
    \item \textbf{Ours (Two-Stage)}: The complete two-stage pipeline, where the first stage performs SFT financial knowledge injection, and the second stage combines GRPO and SFT for reasoning enhancement and knowledge refinement.
\end{itemize}

All configurations are evaluated on the same benchmark datasets: \textbf{FinEval 1.0}, \textbf{FinanceIQ}, \textbf{Finova}, \textbf{MATH-500}, and \textbf{GPQA-diamond}.

\begin{table}[ht]
\centering
\resizebox{\textwidth}{!}{
\begin{tabular}{lccc|cc|c}
\toprule
\textbf{Training Strategy} & \multicolumn{3}{c}{\textbf{Financial}} & \multicolumn{2}{c}{\textbf{General}} & \textbf{Average} \\
\cmidrule(lr){2-4} \cmidrule(lr){5-6}
& \textbf{FinEval 1.0} & \textbf{FinanceIQ} & \textbf{Finova} & \textbf{MATH} & \textbf{GPQA} & \textbf{All Datasets} \\
\midrule
Single-Stage (SFT Only) & 84.19 & 82.32 & 62.87 & \textbf{\underline{94.20}} & 58.59 & 76.43 \\
\textbf{Ours (Two-Stage)} & \textbf{\underline{85.09}} & \textbf{\underline{84.24}} & \textbf{\underline{63.56}} & 93.40 & \textbf{\underline{59.60}} & \textbf{\underline{77.18}} \\
\midrule
Base Model (Qwen3-8B)    & 76.27 & 73.06 & 54.45 & 93.80 & 59.60 & 71.44 \\
\bottomrule
\end{tabular}}
\caption{Performance comparison across different training strategies. Datasets are grouped into \textit{Financial} and \textit{General} categories. Note that MATH refers to MATH-500 and GPQA refers to GPQA-diamond.  \textbf{\underline{Bold}} indicates the best performance among all training configurations.}
\label{tab:twostage_ablation}
\end{table}

\textbf{Key Findings:}
\begin{itemize}
    \item The first-stage SFT brings significant improvement
    \item The second-stage GRPO+SFT further advances the performance upper bound.
    \item The improvement is most pronounced on financial specialized tasks validating the effectiveness of domain-specific training
\end{itemize}

\paragraph{Discussion}
These experiments collectively validate the effectiveness of our two-stage training strategy:

\begin{enumerate}
    \item \textbf{Stage-wise improvement}: Two-stage training brings significant performance gains compared to single-stage training
    \item \textbf{Component synergy}: The combination of GRPO and SFT in the second stage produces optimal results
    \item \textbf{Efficiency gains}: Achieves dual optimization of performance and efficiency while maintaining high performance
\end{enumerate}

These results demonstrate that our two-stage training strategy achieves dual optimization of performance and efficiency through progressive knowledge injection and reasoning enhancement.

\section{Conclusion}
\label{sec:conclusion}

In this work, we introduce \textbf{Agentar-Fin-R1}, a family of \textit{specialized}, \textit{efficient}, and \textit{reasoning-enhanced} financial large language models that systematically addresses fundamental challenges in domain-specific language model development. Our comprehensive approach demonstrates three principal contributions validated through rigorous experimental analysis.

\textbf{Domain Expertise}: Agentar-Fin-R1 achieves state-of-the-art performance across comprehensive financial benchmarks, establishing new performance standards that consistently surpass existing approaches. The specialization manifests most prominently in three critical dimensions: (1) \textit{Agent Capabilities} -- demonstrating sophisticated multi-step reasoning, tool integration, and complex task decomposition in financial workflows; (2) \textit{Hallucination Mitigation} -- maintaining factual precision and epistemic reliability in high-stakes financial decision-making contexts; and (3) \textit{Safety and Regulatory Compliance} -- ensuring adherence to stringent regulatory frameworks and ethical guidelines. Our data construction methodology leverages authenticated, high-fidelity sources through a principled synthesis framework that preserves data provenance and maintains quality assurance throughout the pipeline.

\textbf{Training Efficiency}: We propose a novel weighted training framework incorporating difficulty-aware sample estimation and two-stage optimization strategies that systematically maximizes data utilization efficiency while achieving superior convergence properties. This methodology enables targeted optimization across heterogeneous financial task distributions without incurring the computational penalties typically associated with domain specialization, thereby establishing a new paradigm for efficient domain adaptation. Our sophisticated financial label system further enhances training efficiency by enabling strategic sample selection and curriculum learning approaches that optimize resource allocation.

\textbf{Advanced Reasoning}: Our models demonstrate robust reasoning capabilities that extend beyond specialized financial tasks, maintaining competitive performance on general-domain reasoning benchmarks while achieving domain expertise. This dual competency validates our approach's ability to circumvent catastrophic forgetting while successfully acquiring specialized knowledge representations.

We present \textbf{Finova}, a meticulously designed evaluation suite specifically engineered to comprehensively assess real-world deployment capabilities. 

This research establishes foundational principles for developing trustworthy, efficient, and capable domain-specific language models with broad implications for specialized AI applications. The methodological contributions presented herein extend beyond financial domains to other mission-critical applications where specialization, computational efficiency, and reasoning fidelity are paramount. Future research directions include real-time adaptation mechanisms for dynamic environments and cross-domain generalization of our proposed frameworks.

\clearpage
\bibliographystyle{plainnat}
\bibliography{main}

\newpage
\section*{Appendix}

\begin{figure*}[htbp]
    \centering
    \includegraphics[width=0.95\linewidth]{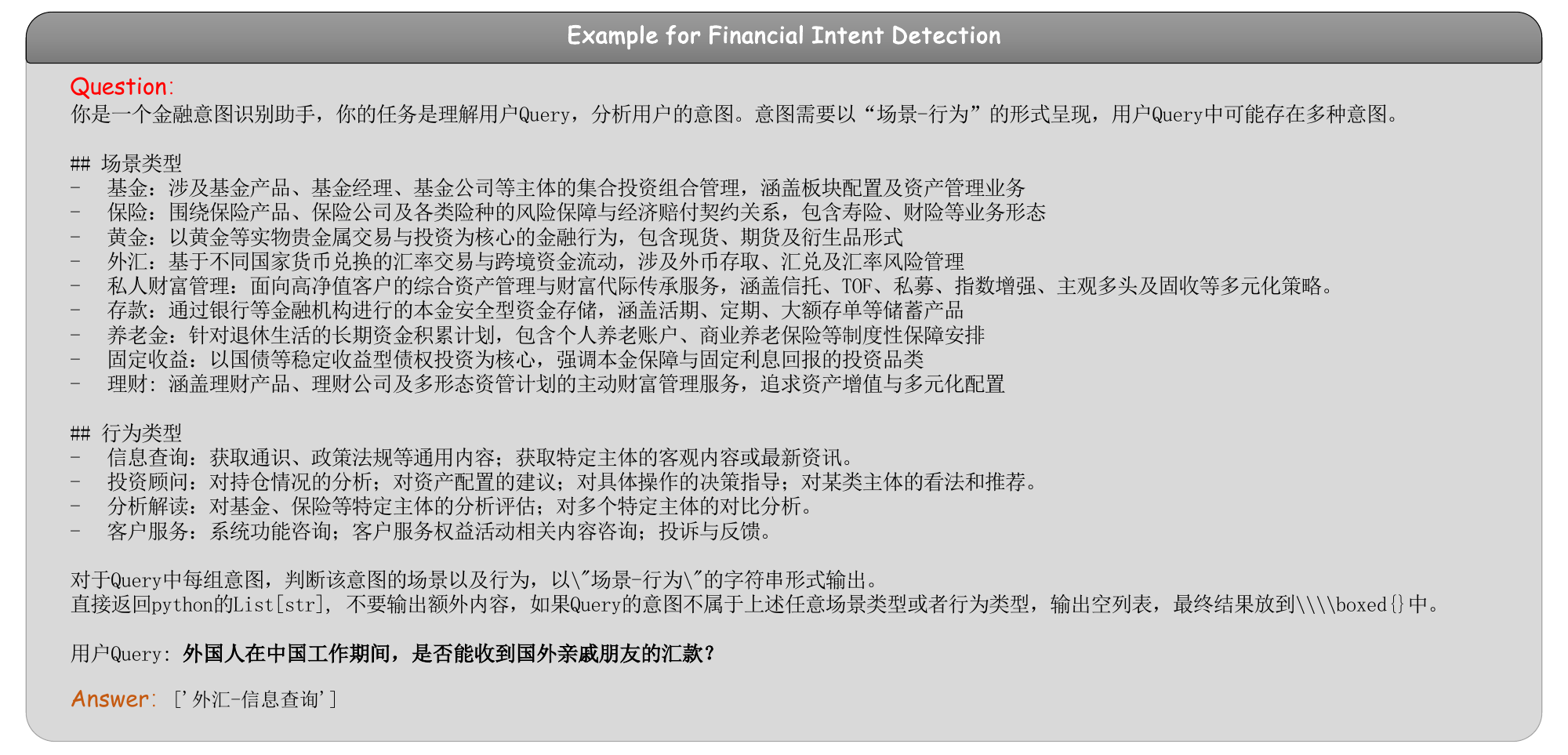}
    \caption{An example for Financial Intent Detection of Finova.}
    \label{fig:intent}
\end{figure*}

\begin{figure*}[htbp]
    \centering
    \includegraphics[width=0.95\linewidth]{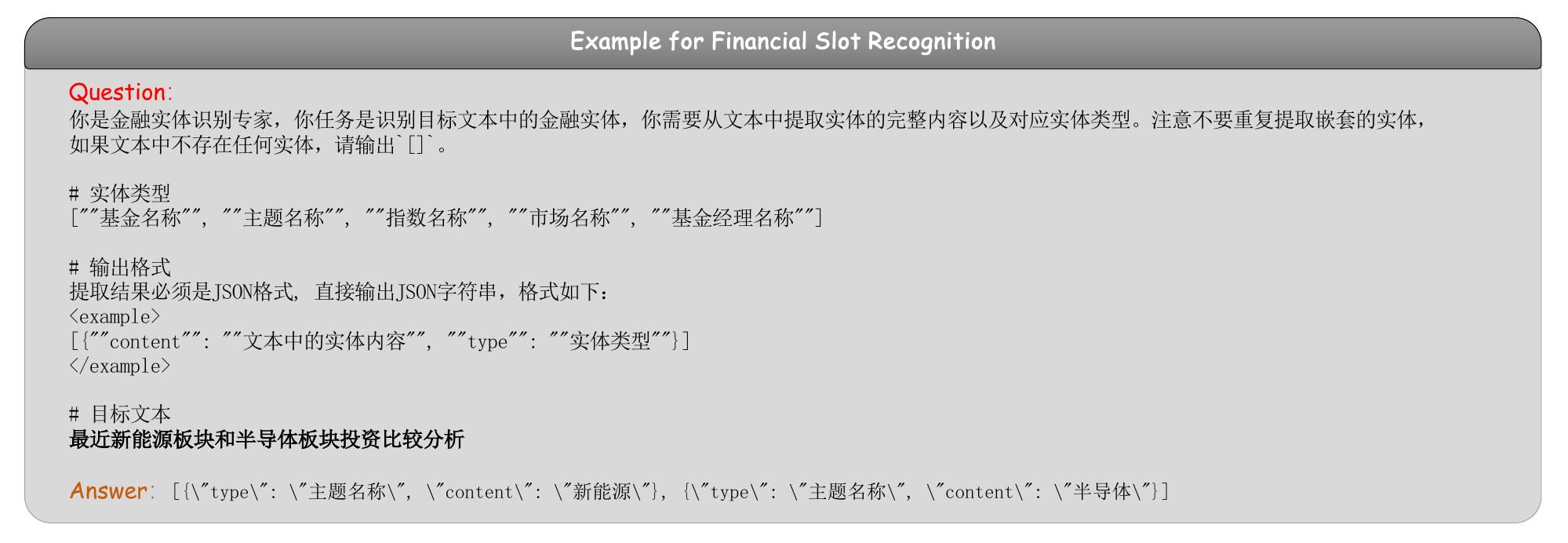}
    \caption{An example for Financial Slot Recognition of Finova.}
    \label{fig:slotting}
\end{figure*}

\begin{figure*}[htbp]
    \centering
    \includegraphics[width=0.95\linewidth]{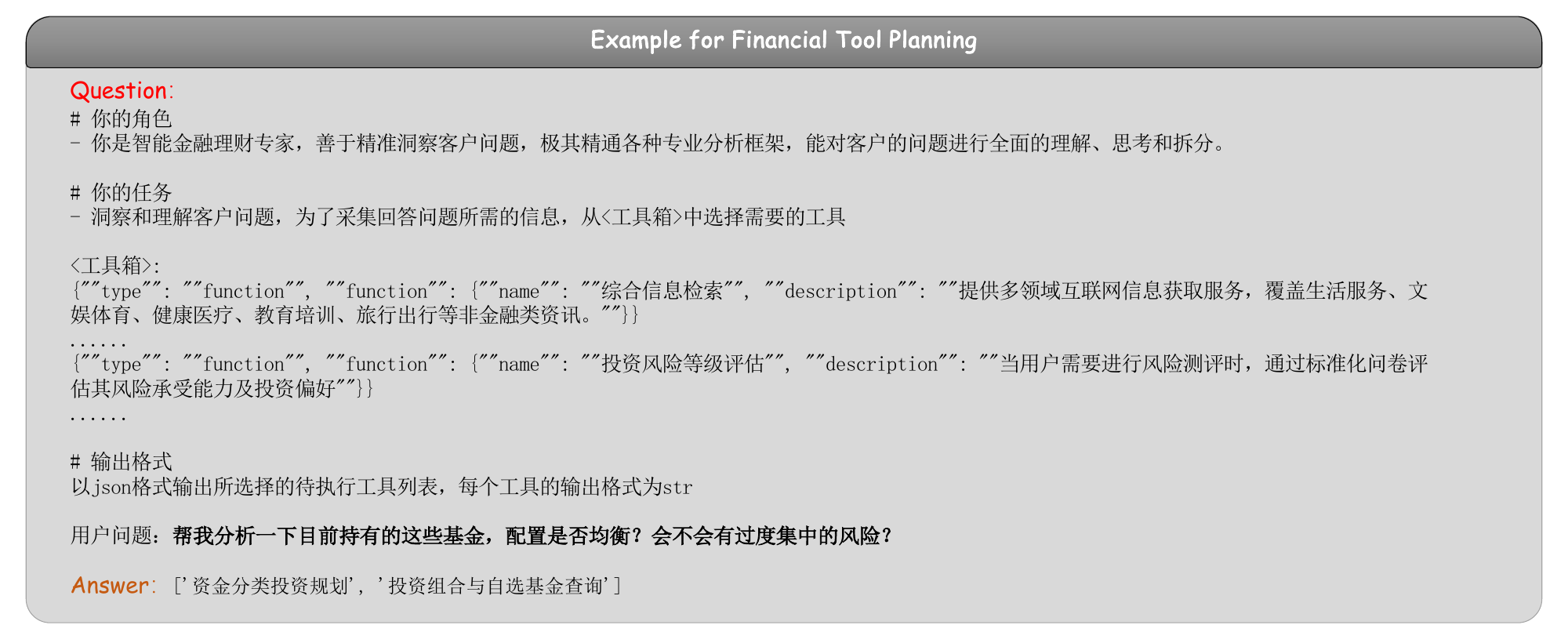}
    \caption{An example for Financial Tool Planning of Finova..}
    \label{fig:tool}
\end{figure*}

\begin{figure*}[htbp]
    \centering
    \includegraphics[width=0.95\linewidth]{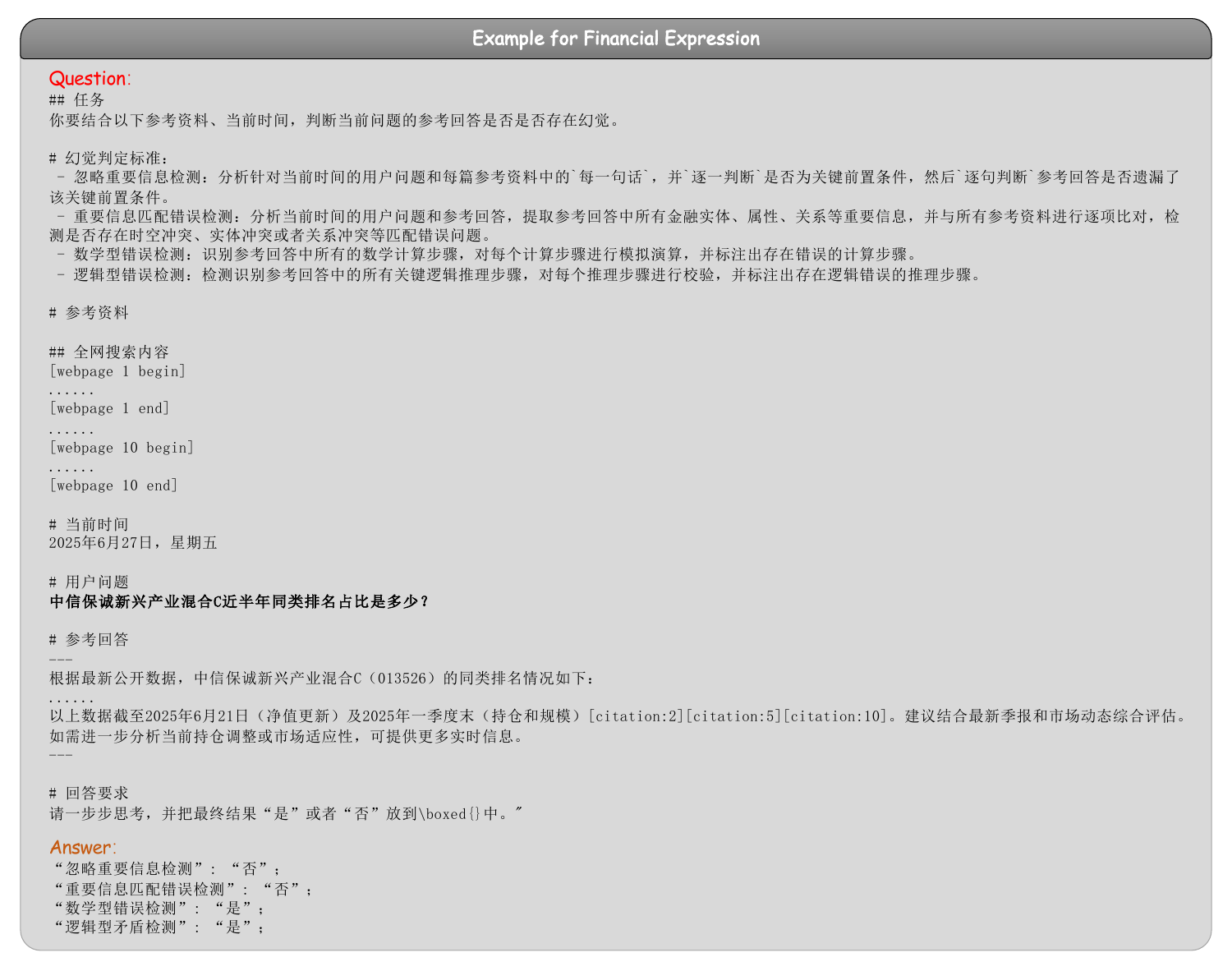}
    \caption{An example for Financial Expression Generation of Finova.}
    \label{fig:expression}
\end{figure*}
\end{document}